%% file: main_arxiv.tex
\documentclass[11pt]{article}

\usepackage[preprint]{acl}

\usepackage{times}
\usepackage{latexsym}
\usepackage{amsmath}
\usepackage{amssymb}
\usepackage{mathtools}
\usepackage{bm}
\usepackage{dsfont}
\usepackage{array}
\usepackage{tabularx}
\usepackage{booktabs}
\usepackage{multirow}
\usepackage[table]{xcolor}
\usepackage{subcaption}
\usepackage{tikz}
\usetikzlibrary{calc}
\usepackage{cleveref}
\usepackage[most]{tcolorbox}

\newtcolorbox{promptbox}[1][]{
  colback=gray!5,
  colframe=gray!50,
  boxrule=0.5pt,
  arc=2pt,
  fontupper=\small\ttfamily,
  title=#1,
  coltitle=black,
  fonttitle=\small\bfseries
}

\usepackage[T1]{fontenc}

\usepackage[utf8]{inputenc}

\usepackage{microtype}

\usepackage{inconsolata}

\usepackage{graphicx}
\usepackage{tikz}
\usepackage{pgfplots}
\pgfplotsset{compat=1.18}
\usetikzlibrary{arrows.meta}

\title{Error-Type-Aware Loss Reweighting for Robust Named Entity Recognition with Noisy LLM Labels}

\author{Elena Merdjanovska\textsuperscript{1,2}, Jonas Golde\textsuperscript{1} and Alan Akbik\textsuperscript{1,2} \\ 
 \textsuperscript{1}Humboldt-Universität zu Berlin \\ \textsuperscript{2}Science of Intelligence \\
 \texttt{\{elena.merdjanovska, jonas.max.golde.1, alan.akbik\}@hu-berlin.de}}

\begin{document}
\maketitle

\begin{abstract}
Large language models are increasingly used to annotate datasets for training smaller, task-specialized models such as named entity recognition. While this method yields effective models, it assumes that the synthetic dataset is correctly annotated. In this work, we find that \textit{(i)} current fine-tuning processes simply ignore LLM-introduced annotation noise, resulting in degraded performance and \textit{(ii)} existing noise-robust losses are not transferable to sequence labeling because annotation noise in named entity recognition is heterogeneous: for example, missing mentions and type errors affect the training signal in different ways. 
Treating all noisy tokens equally in noise-robust losses and applying a single reweighing criterion for all may therefore remove useful supervision or reinforce incorrect labels. To address this limitation, we propose error-type-aware loss reweighting for NER, which introduces separate reweighing rules for different types of potentially erroneous tokens. 
Our approach is simple and efficient, does not require additional training resources, and improves F1 by 0.8 - 2.0 percentage points on dataset-level average for noise levels between 15\% and 40\%, with a maximum improvement of 4.6 percentage points with 24.1\% noise on Wikigold.\footnote{We release the code in  \url{https://github.com/elenamer/error-type-aware-losses}} 


\end{abstract}

\section{Introduction}

Recently, large language models (LLMs) are used as an alternative to manual annotation \citep{ding-etal-2023-gpt,pavlovic-poesio-2024-effectiveness} as they reduce annotation costs and generate reliable supervision for large, previously unlabeled datasets \citep{tan-etal-2024-large}. These annotations can be used to fine-tune smaller models for domain-specific tasks, including named entity recognition (NER). Compared with general-purpose LLMs, such smaller fine-tuned models provide lower latency, reduced memory use, and cheaper deployment \citep{bogdanov-etal-2024-nuner,zaratiana-etal-2024-gliner,golde-etal-2026-finerweb}. 

\input{figures/introduction_figure}

Despite these benefits, LLM-generated annotations are often incomplete or incorrect. For example, when looking at \Cref{fig:teaser-ceiling}, we observe that prompting achieves 66.2 F1 on the gold-annotated test split of OntoNotes. At the same time, when fine-tuning on the LLM-annotated train split and evaluate on the corresponding clean test split, we observe across all four datasets in \Cref{fig:teaser-ceiling} that the fine-tuned F1 stays within 3 points of the prompting approach. Thus, the annotation quality creates a practical ceiling on downstream performance performance and LLM-generated annotations directly transfers its errors to the smaller model.


Noise-robust learning approaches try mitigating the problem of erroneous labels in the training process. Existing approaches include sample selection, label correction, loss reweighting, regularization, and multi-network learning \citep{frenay2013classification,han-etal-2018-colearning,song2023learning}. For example, loss reweighting aims to assign lower weight to incorrect annotations when calculating the total loss \cite{liu2016importance,pmlr-v97-arazo19a} which prevents noisy examples from dominating the optimization process. 

However, these approaches are primarily designed for classical classification problems. NER is a structured prediction problem in which one jointly models the detection entity types and span boundaries. Thus, the errors in NER datasets are of heterogeneous nature: \textit{(i)} the LLM may miss an entity mention\footnote{We use the following names for each error type throughout the paper: \textit{(i) missing}, \textit{(ii) false positive}, \textit{(iii) type} and \textit{(iv) partial}.}, \textit{(ii)} hallucinate one that is not present, \textit{(iii)} assign the wrong type or \textit{(iv)} annotate the mention boundary incorrectly \citep{merdjanovska-etal-2024-noisebench}. Further, entity annotations are sparse and the majority of tokens is assigned the non-entity label \textit{O}, resulting in a severe class imbalance. Treating all noisy tokens equally may therefore remove useful supervision or reinforce incorrect labels. Most prior NER-specific work on noisy supervision has focused on distant supervision rather than LLM-generated annotations which exhibit their very own error patterns \cite{zhang-etal-2025-dynclean,li-etal-2025-examine}. 

To address this limitation, we propose error-type-aware loss reweighting for NER. Our idea is simple: we assume that most LLM annotations are correct and worth learning from, thus our goal becomes identifying and ignoring incorrectly labeled tokens from the loss. Since we assume that annotations are correct, we can treat disagreement between the models' prediction and actual annotation as the signal that a token may be mislabeled by using the models' confidence. We can make this distinction for all of the previous error types such that we can adapt what we want to mask, depending on the noise occurring in the LLM annotated data.

We summarize our contributions as follows:
\begin{enumerate}
\item We analyze the main error types in LLM-generated NER annotations and find that errors differ greatly across datasets and prompting approaches, even with the same model. 
\item We propose six error-type-aware reweighting loss variants, that mask-out potentially unreliable tokens so they do not negatively impact the training process. 
\item We evaluate the losses across four datasets, three LLM annotators and two models against standard fine-tuning and representative noise-robust baselines. 
\item Our results show that the proposed error-type-aware reweighting is effective, with per-dataset F1 improvements over standard cross-entropy between 0.8 and 2.0 percentage points with \texttt{DistilBERT} and 1.2 - 1.6 with \texttt{XLM-RoBERTa (large)}. 
\item We show that error-type-aware reweighting is more effective than global token reweighting, and find that $L_{\text{missing}}$ that targets missing annotations is the best-performing loss overall.
\end{enumerate}

\section{Related Work}

\noindent\textbf{Noisy Labels in NER.}~Noise-robust NER has been studied extensively for distant supervision. BOND \citep{bond2021} uses self-training to replace distant labels with model-generated pseudo-labels, while RoSTER \citep{meng-etal-2021-distantly} combines noisy-label removal and self-training. NEEDLE \citep{jiang-etal-2021-named} similarly incorporates a noise-aware loss and self-training under weak supervision. 

Many approaches identify unreliable tokens using model confidence or training dynamics, filtering for positive and negative mention samples separately \citep{zhang-etal-2025-dynclean, debiased2023}. However, most of these studies, use additional training resources, for example \citet{li-etal-2025-examine} use many trained models as voters to estimate label reliability. \citet{liu-etal-2021-noisy-labeled} use multiple training iterations in their confidence estimation method, while the approaches by \cite{zhang-etal-2025-dynclean} and \citet{merdjanovska-akbik-2025-token} require final re-training using the cleaned dataset. Self-Cleaning \cite{chu-etal-2024-self}, trains a discriminator on a small set of clean instances and uses its predictions to reweigh tokens affected by boundary and type errors separately. Like these approaches, ours filters tokens by error type, but does so directly in the loss, without additional training stages or a clean validation subset.

\citet{zhu-etal-2023-weaker} show that distant- and weak-supervision methods relying on clean development data degrade substantially when that data is noisy, and that a small clean set is better spent on fine-tuning than on model selection. We therefore adopt the more realistic setting in which no clean data is available.

\noindent\textbf{LLMs as Annotators.}~Although most research on noisy NER focuses on weak or distant supervision, recent benchmarks found that increasngly different annotation sources introduce different noise types into the training process, including LLM-generated labels \cite{merdjanovska-etal-2024-noisebench,debiased2023}. In recent years, LLM-generated labels have become increasingly common across many tasks, including NER. Several zero-shot and few-shot prompting approaches have been proposed, such as GPT-NER \citep{wang-etal-2025-gpt}, DiZiNER \citep{kim-yoon-2026-diziner}, and EvoPrompt \citep{tong-etal-2025-evoprompt}. DiZiNER and EvoPrompt both use prompt-optimization pipelines with iterative refinement. Hybrid annotation pipelines that combine human and LLM labels have also been shown to recover performance under partial annotation loss or limited annotation budgets \cite{naraki2024augmentingnerdatasetsllms}. These studies focus on improving annotation quality rather than on training robustly with the resulting labels, which remain noisy. We address this gap by treating LLM-generated labels as a primary source of noisy supervision.

\section{Method}

\subsection{Problem Setup}

Let $x_i$ denote a token instance and let $\tilde{y}_i$ be its observed noisy label. We assume the training labels may come from a noisy annotation process, and that their corruption probability is \emph{instance dependent} \cite{beigman-beigman-klebanov-2009-learning,xia2020partdependentlabelnoiseinstancedependent}. In particular, the probability that $\tilde{y}_i$ differs from the latent clean label varies with the instance and the observed label. The goal is to train an NER model $f_\theta$ that performs well on a clean test set despite noisy supervision.

We consider token-level NER, where each token in a sequence is assigned a label from a set $\mathcal{Y} = \{O\} \cup \mathcal{E}$, with $O$ denoting the non-entity (outside) class and $\mathcal{E}$ denoting the set of entity types with BIO tagging. The outside class $O$ dominates the label distribution of tokens in NER.

Let $\ell(f_\theta(x_i), \tilde{y}_i)$ denote the base loss for instance $i$, token-level categorical cross-entropy. 
Our approach introduces a binary reliability mask for loss reweighting $w_i \in \{0,1\}$ and the total loss is:
\begin{equation}
\mathcal{L} = \frac{1}{N}\sum_{i=1}^{N} w_i \, \ell(f_\theta(x_i),\, \tilde{y}_i).
\end{equation}
This is an importance-weighted empirical risk estimator \cite{liu2016importance}, where $w_i$ is derived from a hard decision about the latent label-quality state of each token.

\subsection{Latent Clean Indicator}

Let $z_i \in \{0,1\}$ be a latent indicator, where $z_i=1$ means that the observed label $\tilde{y}_i$ is clean and $z_i=0$ means that it is corrupted. We do not observe $z_i$ directly. From a confidence score $s_i = \max_c \, p_\theta(c \mid x_i)$, a fitted mixture model (Section~\ref{sec:threshold}) yields the posterior
\begin{equation}
q_i = P(z_i{=}1 \mid x_i, \tilde{y}_i, \hat{y}_i, s_i),
\end{equation}
where $\hat{y}_i = \arg\max_c \, p_\theta(c \mid x_i)$ is the model's current prediction. We then take its \emph{maximum a posteriori (MAP) estimate}
\begin{equation}
\hat{z}_i = \mathds{1}\!\left[q_i > \tfrac{1}{2}\right] =
\begin{cases}
1, & q_i > \tfrac{1}{2},\\
0, & \text{otherwise.}
\end{cases}
\end{equation}
where $\hat z_i$ is the most likely label-quality state under the fitted mixture model. 

\subsection{Adaptive Threshold Estimation}
\label{sec:threshold}

To avoid reliance on clean validation labels, we estimate the mixture model underlying $q_i$ from the training data itself. We fit a two-component Beta mixture (BMM) to the score distribution and interpret one component as likely clean and the other as likely noisy \cite{pmlr-v97-arazo19a,li2020dividemix}. The MAP boundary $\tau$ is the point at which posterior mass is split evenly between the two components,
\begin{equation}
P(z=1 \mid s=\tau) = P(z=0 \mid s=\tau),
\end{equation}
and the MAP state reduces to a simple threshold comparison:
\begin{equation}
\hat{z}_i = \mathds{1}[s_i > \tau].
\end{equation}
That is, once the mixture components are estimated and $\tau$ is set at the equal-posterior boundary, a token is classified as clean if and only if its confidence $s_i$ exceeds $\tau$.

In practice, we start training with a fixed $\tau$ value, update it after the first few epochs, and refresh it two more times during training at equal intervals.

\subsection{Error-Type-Aware Hard Reweighing}
\label{sec:hard-masking}

We convert the MAP state $\hat z_i$ into a loss weight only when the model disagrees with the observed label \cite{debiased2023}. We use
\begin{equation}
w_i =
\begin{cases}
1, & \text{if } \hat{y}_i = \tilde{y}_i,\\
1, & \text{if } \hat{y}_i \neq \tilde{y}_i \text{ and } \hat{z}_i = 1,\\
0, & \text{if } \hat{y}_i \neq \tilde{y}_i \text{ and } \hat{z}_i = 0,
\end{cases}
\end{equation}
where a weight of $0$ amounts to hard masking, i.e.\ complete exclusion of the token from the loss. 

We introduce separate masking rules for different types of tokens. We partition the disagreeing tokens into error-type-specific sets
:
\begin{align}
\mathcal{I}_{\mathrm{missing}} &= \{i : \tilde{y}_i = O,\; \hat{y}_i \neq O\},\\
\mathcal{I}_{\mathrm{entity}} &= \{i : \tilde{y}_i \neq O,\; \hat{y}_i \neq \tilde{y}_i\},
\end{align}
corresponding to missing-entity and entity errors, respectively.\footnote{It should be noted that according to the categorization of NER errors, tokens from both $\mathcal{I}_{\mathrm{missing}}$ and $\mathcal{I}_{\mathrm{entity}}$ can also be a part of a boundary error. The naming follows from the most common error type associated with these token types.} 

\noindent
As further method variants, we partition the tokens from $\mathcal{I}_{\mathrm{entity}}$ into two more sets:
\begin{align}
\mathcal{I}_{\mathrm{type}} &= \{i : \tilde{y}_i \neq O,\; \hat{y}_i \neq O\},\\
\mathcal{I}_{\mathrm{FP}} &= \{i : \tilde{y}_i \neq O,\; \hat{y}_i = O\},
\end{align}
corresponding to false positive (hallucinated) entities and type errors, respectively.

\noindent
As a baseline we include a single threshold variant that applies one mixture model to all misclassified tokens \citep{meng-etal-2021-distantly}: 
\begin{align}
\mathcal{I}_{\mathrm{all}} &= \{i : \hat{y}_i \neq \tilde{y}_i\},
\end{align}

\noindent
Let $\mathcal{K}$ denote the error-type partitions (one or multiple) used in a given variant, with token sets $\{\mathcal{I}_k\}_{k\in\mathcal{K}}$ \footnote{For example: \ $\mathcal{K}=\{\mathrm{missing}, \mathrm{entity}\}$ or $\mathcal{K}=\{\mathrm{type}\}$}. For each error type $k$, we fit a separate mixture model to the confidence scores of the tokens in $\mathcal{I}_k$, yielding a type-specific posterior $q_i^{(k)}$ and MAP state $\hat{z}_i^{(k)}$.

\noindent
The final binary mask is then
\begin{equation}
w_i = \mathds{1}[\hat{y}_i = \tilde{y}_i]
  + \sum_{k\in\mathcal{K}} \mathds{1}[i \in \mathcal{I}_k]\,\hat{z}_i^{(k)},
\end{equation}
where the first term keeps agreement tokens in the loss, and the second term includes a disagreement token only if the MAP state under its error-type-specific mixture is clean. Otherwise the token is masked entirely ($w_i=0$).

Our final set of losses evaluated in Section \ref{sec:loss-results} is: $L_{\text{all}}$, $L_{\text{missing}}$, $L_{\text{entity}}$, $L_{\text{type}}$, $L_{\text{FP}}$ and $L_{\text{missing, entity}}$.

\section{Experimental Setup}
Across all experiments, we adopt the evaluation protocol of \citet{merdjanovska-etal-2024-noisebench}. We train models on noisy annotations but evaluate on a clean test split with gold labels. This isolates the effect of annotation noise on the learned model, since the test signal is unaffected by the noise process. Further, we assume no clean data is available at any stage, so we use noisy validation split for model selection. For each dataset we consider several annotation sources spanning a range of noise levels and error profiles.

\subsection{Datasets}

We evaluate on four NER datasets: NoiseBench \cite{merdjanovska-etal-2024-noisebench} --- based on CoNLL03 \cite{tjong-kim-sang-de-meulder-2003-introduction, rucker-akbik-2023-cleanconll}, OntoNotes \citep{weischedel2013ontonotes}, Wikigold \citep{balasuriya-etal-2009-named}, and BC5CDR \citep{wei2016assessing}. We selected these datasets to cover different label set sizes, with NoiseBench and Wikigold both including 4 types (PER, LOC, ORG and MISC), BC5CDR covering 2 types (DISEASE and CHEMICAL) and OntoNotes covering 18 types.

In addition to the LLM-annotated versions, used in our main experiments, we also evaluate on distant supervision labels \citep{bond2021, shang2018learning}. Additionally, NoiseBench has 5 further noisy variants. See Appendix \ref{appendix:datasets} for an overview of the extended noise sources and further dataset details.

\subsection{LLM Annotation}

Our primary noise source are LLM annotations. We use two models: \texttt{gpt-oss-120b} and \texttt{Qwen3-235B-A22B-Instruct-2507}, and four prompting approaches: basic prompting, schema prompting, DiZiNER \citep{kim-yoon-2026-diziner} and EvoPrompt \citep{tong-etal-2025-evoprompt}, to get a diverse set of noisy annotations\footnote{We do not aim to assess the effectiveness of these approaches. We adapted them with the goal of obtaining varying performance levels, rather than achieving the best possible prompting for NER.}. The prompts and details are given in Appendix \ref{appendix:llm-annotation}. The final selected label variants, three per dataset, are shown in Table \ref{tab:noise-levels-llm-labels}. 

\subsection{Baselines}

We compare against standard categorical cross-entropy (CE) and the following noise-robust loss baselines: generalized cross-entropy (GCE) \cite{zhang2018generalized, meng-etal-2021-distantly}, focal loss \citep{lin2018focallossdenseobject}, BMM bootstrap loss --- $\text{BMM}_{\text{b}}$. \citep{pmlr-v97-arazo19a}, corrected NLL --- $\text{NLL}_{\text{c}}$ \citep{jiang-etal-2021-named}. 

\section{Results}
In this section, we give an overview of the LLM annotation errors, evaluate the performance of our error-type-aware losses and compare them to existing noise-robust losses. Further, we compare them to ideal masking upper bounds, and analyze their masking dynamics.

\subsection{LLM Annotation Errors}
We first want to understand the extent of LLM-introduced annotation noise for NER. To do so, we generate three LLM annotations for each dataset, as listed in Table \ref{tab:noise-levels-llm-labels}. This allows us to evaluate our methods across varying noise levels while consistently reflecting LLM annotation noise. For example, on OntoNotes, the approaches range from 60.7 to 70.8 F1 on the training set, while on NoiseBench, a CoNLL subset, scores range from 75.1 to 85.0.
\input{tables/exp1_datasets_overview_llm}

\noindent \textbf{Error types.} The annotations exhibit different error profiles, as shown in the \textit{\%Errors} columns of Table \ref{tab:noise-levels-llm-labels}. For example, in BC5CDR that has only two entity types, most errors across all LLM label variants are \textit{missing} mentions (\textit{FN}), while \textit{type} errors are rare. In contrast, different prompts on NoiseBench produce substantially different error profiles, often with many \textit{type} errors. The best-performing label variants generally have the most balanced profiles: in the highest-quality LLM labels for OntoNotes, NoiseBench, and Wikigold, no single error type accounts for more than 40\% of erroneous mentions.






\input{tables/exp2_table_distilbert}

\subsection{Evaluation of Error-Type-Aware Losses}
\label{sec:loss-results}

In this section, we evaluate the effectiveness of our proposed error-type-aware losses. We show a comparison across all datasets and LLM-generated label variants with \texttt{DistilBERT} in Table \ref{tab:distilbert-base-uncased-test-f1-wide}. 

\noindent \textbf{Best-performing losses.} $L_{\text{missing}}$ is the best loss on all OntoNotes and BC5CDR label variants, improving over cross-entropy by approximately 1 F1 point on average. $L_{\text{entity}}$ performs best on NoiseBench, with its largest gain on \texttt{GPT-OSS Basic}, where F1 increases from 68.9 to 72.5. Results on WikiGold are less consistent, with different losses performing best across label variants. It is also the only dataset on which the more error-specific $L_{\text{FP}}$ variant performs best on average. Its largest improvement occurs for \texttt{GPT-OSS EvoPrompt}, from 60.2 to 64.8. 

\noindent \textbf{Loss performance aligns with the error types.} Wikigold's \texttt{GPT-OSS EvoPrompt} also has the highest proportion of false-positive (FP) errors (Table \ref{tab:noise-levels-llm-labels}), suggesting that the effectiveness of an error-type-aware loss depends on the underlying error distribution. More generally, in 9 of the 12 label variants, the best-performing loss targets the most frequent error category, as indicated by the column colors in Table \ref{tab:distilbert-base-uncased-test-f1-wide}.

\noindent \textbf{Benefits and limits of error-specific masking.} 
Separating tokens according to their observed labels, as in $L_{\text{missing}}$ and $L_{\text{entity}}$, is beneficial and significantly outperforms the global variant $L_{\text{all}}$ across all datasets and noise types. However, more specific masking through further partitioning of $L_{\text{entity}}$ according to the predicted label into $L_{\text{FP}}$ and $L_{\text{type}}$ provides no consistent advantage. For example, $L_{\text{type}}$ performs poorly even on the NoiseBench \texttt{GPT-OSS EvoPrompt} labels, where type errors account for more than 50\% of the noise. In general, the broader $L_{\text{entity}}$ variant is more effective for both type errors and false positives.

\input{figures/lineplot_noisebench_F1s}

\noindent \textbf{Comparison to related work losses.}
We compare our loss variants with existing noise-robust losses in Table \ref{tab:distilbert-base-uncased-test-f1-dataset-averages-wide}. None of the existing losses outperform our variants on any dataset. This includes general NER objectives such as Focal Loss, general noise-robust objectives such as GCE and BMM Bootstrap, and NER-specific noise-robust objectives such as Corrected NLL. Although these losses occasionally outperform cross-entropy (for example, Focal Loss on Wikigold) their overall results remain close to the baseline.

\noindent \textbf{Findings extend to another model.} We observe similar result patterns with \texttt{XLM-RoBERTa (large)} (Table \ref{tab:xlm-roberta-large-test-f1-wide} and Table \ref{tab:xlm-roberta-large-test-f1-dataset-averages-wide} in Appendix \ref{appendix:xlm-roberta_results}). The main difference is on Wikigold, where $L_{\text{missing}}$ performs best with \texttt{XLM-RoBERTa-large}, whereas $L_{\text{entity}}$ performs best with \texttt{DistilBERT}. However, the \texttt{DistilBERT} results for $L_{\text{missing}}$, $L_{\text{entity}}$, and $L_{\text{FP}}$ are very close, suggesting that this difference is small.

\input{tables/exp2_comparison_to_other_losses_distilbert}

\subsection{Upper Bounds for Loss Masking}

We next examine the general potential of loss masking using two oracle-based approaches:
\begin{itemize}
\item \textit{Ideal Masking} --- masks all erroneous tokens, identified by comparing the BIO labels in the noisy and clean label variants. The mask remains fixed throughout training.
\item \textit{Ideal $L_{\text{all}}$} --- masks an erroneous token only when it is misclassified by the model. The mask therefore changes across epochs with the model's predictions.
\end{itemize}

\noindent \textbf{High potential of masking approaches.}
Ideal Masking achieves test F1 scores close to those obtained by fine-tuning on clean data (Figure \ref{fig:F1s_lineplot} on NoiseBench), demonstrating that excluding erroneous tokens from training has the potential to recover almost clean-training performance. Ideal $L_{\text{all}}$ also significantly improves performance under label noise, although by a smaller margin.

The two approaches represent different performance limits. Ideal Masking provides an absolute limit because it removes all erroneous tokens, including those the model might memorize easily. Such tokens are difficult to identify from confidence scores \citep{merdjanovska-akbik-2025-token}. Ideal $L_{\text{all}}$ is more realistic because it masks only erroneous tokens that remain misclassified \cite{chong-etal-2022-detecting, debiased2023}. As it is also conceptually closer to our proposed losses, we treat Ideal $L_{\text{all}}$ as a method-level upper bound.

\noindent \textbf{Error-aware masking narrows the gap to ideal.} Despite the strong performance of Ideal $L_{\text{all}}$, its global practical counterpart $L_{\text{all}}$ performs poorly across all noise types and is often worse than the CE baseline (Figure \ref{fig:F1s_lineplot}). This method applies a single confidence threshold to both O and entity tokens, while $L_{\text{missing}}$ and $L_{\text{entity}}$ treat each token type separately. $L_{\text{missing}}$ performs well for \texttt{Crowd++}, \texttt{Distant}, and \texttt{Crowd} noise, whereas $L_{\text{entity}}$ outperforms the baseline for \texttt{GPT-OSS-Schema}, \texttt{GPT-OSS Basic}, \texttt{Weak}, and \texttt{GPT3.5} noise. This again shows that masking the token groups separately is considerably more effective.  



\subsection{Masking Dynamics}
We analyse the behavior of our proposed $L_{\text{missing}}$ through the training iterations. 

\noindent \textbf{Adaptive thresholding results in consistent masking patterns.}
In Figure \ref{fig:pu-loss-thresholds} we see how the threshold of $L_{\text{missing}}$ masking dynamically adapts for different noise levels. It generally increases in later epochs, although it can decrease in some settings, such as \texttt{Weak} noise. Despite these differences, the number of masked tokens follows a consistent pattern across noise levels (Figure \ref{fig:pu-masked-tokens}): it is initially high and gradually decreases. This is expected since more tokens are misclassified early in training and recovering noisy samples is more effective during early-learning \citep{pmlr-v70-arpit17a}.

Higher noise levels also lead to more masked tokens, as intended. Thus, although the threshold trajectories differ across label variants, they produce similar masking dynamics. This suggests that the adaptive thresholding mechanism is effective
.
\input{figures/thresholds_pu_loss}

\input{figures/combined_losses_plots}

\subsection{Why Combined Masking Underperforms}

The results for both models confirm that practical $L_{\text{all}}$ is the weakest method and performs substantially worse than CE. A likely explanation is that misclassified O and non-O tokens have different confidence distributions \cite{merdjanovska-akbik-2025-token}. Because most tokens in NER belong to the \textit{O} class, applying a shared threshold to both groups ignores the strong class imbalance and leads to incorrect filtering --- with many correct tokens being masked and incorrect tokens still learned. 

\noindent \textbf{Combined masking filters out too much useful information.} The other combined approach, $L_{\text{missing, entity}}$, also underperforms despite applying separate thresholds to the two token groups. Figure \ref{fig:training-dynamics} compares the dynamics of loss, development F1 and number of masked tokens of the different losses. As expected, $L_{\text{missing, entity}}$ mask over twice more tokens than $L_{\text{missing}}$ or $L_{\text{entity}}$ combined 
, suggesting that its poor performance results from the total amount of information removed.
This interpretation is supported by the training dynamics. Under $L_{\text{missing, entity}}$, the training loss decreases sharply, while F1 on the noisy development set fails to improve. The combined approach therefore appears too aggressive: by masking both token groups simultaneously, it removes too much training signal for the model to learn the task effectively.

\noindent \textbf{Practical recommendation.} We therefore conclude that each token type should be addressed separately rather than attempting to correct all errors simultaneously. In practice, the loss should be selected according to the expected dominant error type: $L_{\text{missing}}$ for missing (false negatives) and $L_{\text{entity}}$ for false positives and type errors. When neither error type is clearly dominant, $L_{\text{missing}}$ is the safer default because it provides the largest average improvement over standard cross-entropy across datasets and LLM-generated label variants.











\section{Conclusion}
This paper addresses the important issue of fine-tuning under noisy LLM-generated supervision. As LLMs are increasingly adopted for annotation, particularly in information extraction tasks like NER, their variable annotation quality can substantially affect downstream model performance. 

We propose a label noise-robust loss reweighing approach, which targets different types of NER errors separately. To capture a broad range of noise levels and error profiles, we evaluate our approach on four datasets, each with three LLM-generated label variants. Our results show that targeted error-type-aware masking is more effective than uniform masking across all error types. We improve the test F1 scores by between 0.8 and 2.0 percentage point per dataset. Our method is efficient and can be applied directly into a single fine-tuning run without additional resources, making it suitable and practical for fine-tuning small domain-specific NER models on LLM-annotated data.

\section*{Limitations}
We find that noise robust loss functions are most effective when aligning them with the error distribution of the annotation model and we did not find a unified error mode. This distribution can not be known in advance since measuring it requires a small set of clean labels. Further, we did not find a single unified error mode and combined masking approaches yield degraded performance across the benchmarks investigated.

Further, our method assumes that the error types can be explained with two Beta components and that the resulting posterior is monotone in score which we both cannot guarantee. For datasets having fewer examples, less classes, or different distributions, out method may find unreliable threshold and thus yields degraded performance.

Our experiments are limited to two encoder models (\texttt{DistilBERT} and \texttt{XLM-RoBERTa}) and evaluate only on English-data and we do not investigate whether our approach transfers for example to autoregressive models or low-resource languages.

At last, our approaches are evaluated on token-level NER, without considering span-level classification. Span-level approaches are very popular with cross-encoder and bi-encoder generalist models, and often trained using LLM-generated data.

\section*{Acknowledgements}
Elena Merdjanovska and Alan Akbik are funded by the Deutsche Forschungsgemeinschaft (DFG, German Research Foundation) under Germany’s Excellence Strategy – EXC 2002/2 “Science of Intelligence” – project number 390523135. Alan Akbik is further supported by the Deutsche Forschungsgemeinschaft (DFG, German Research Foundation) under Emmy Noether grant “Eidetic Representations of Natural Language” (project number 448414230). Jonas Golde is supported by the Bundesministerium für Bildung und Forschung (BMBF) as part of the project “FewTuRe” (project number 01IS24020). 

\bibliography{custom}

\newpage
\appendix

\section{Experimental Details}
\label{appendix:experimental-details}

\subsection{Datasets}
\label{appendix:datasets}

Table \ref{tab:dataset-sizes} shows the dataset sizes and classes. For NoiseBench, BC5CDR and OntoNotes we used the original train-test splits. For Wikigold which does not have an established split, we split it into train, dev and test 80\% / 10\% / 10\%, based on the document boundaries and the original sentence order. 
We downsample the OntoNotes training set and use a subset consisting of 20\% of the original training set and the full test set. 

For BC5CDR, OntoNotes and Wikigold, we use the Flair \citep{akbik-etal-2019-flair} implementation for the dataset's original (expert) labels. For OntoNotes and Wikigold, we use the BOND distant labels \cite{bond2021}. For BC5CDR, we use the AutoNER distant labels \cite{shang2018learning}. Table \ref{tab:noise-levels-other-labels} shows the noise levels of the distant, weak and crowd noisy variants.

\input{tables/appendix_dataset_sizes}

\input{tables/exp1_datasets_overview_other}

\subsection{LLM Annotation}
\label{appendix:llm-annotation}

We used two main prompts, \textit{Basic} (for all datasets) and a more complex \textit{Schema} prompt (only for Wikigold and NoiseBench). An example \textit{Basic} prompt for NoiseBench is given in Figure \ref{fig:prompt}. 
We also use adapted versions of the prompt refinement algorithms EvoPrompt \citep{tong-etal-2025-evoprompt} and DiZiNER \citep{kim-yoon-2026-diziner}. For both, we used 100 (unlabeled) train sentences, with 2 refinement iterations for EvoPrompt and 4 cycles for DiZiNER. We used two annotator models: \texttt{gpt-oss-120b} and \texttt{Qwen3-235B-A22B-Instruct-2507}. The final prompts for each approach and dataset can be found in our repository.

The various prompts and two models resulted in a total of 4-6 LLM-generated label variants for each dataset. Out of these, we selected 3 representative label variants, ensuring a wide range of noise levels, and always including the highest quality label per dataset.

\begin{figure}[t]
\centering
\begin{promptbox}[Basic Prompt for NoiseBench]
\small\ttfamily
Extract all named entities from the text that match the following categories: PER, ORG, LOC, MISC. Following is the description of each category: \newline
    PER: Named individual people, fictional characters, and person-like deities.\newline
	ORG: Named organizations, institutions, companies, agencies, teams, and media organizations.\newline
	LOC: Named geographical or political locations such as countries, cities, regions, mountains, and waters.\newline
	MISC: Other named entities, including events, works, languages, religions, nationalities, and products.\newline

    CRITICAL OUTPUT INSTRUCTIONS:\newline
    1. Respond ONLY with a valid JSON object. Do not include markdown or explanations.\newline
    2. The JSON keys MUST strictly be: PER, ORG, LOC, MISC.\newline
    3. The value for each key must be a list of objects containing: "mention", "confidence", and "reasoning".\newline

    Example JSON structure:\newline
    {\newline
        "PER": [],\newline
        "ORG": [],\newline
        "LOC": [],\newline
        "MISC": []\newline
    }\newline
    Text to Analyze:\newline
    <SENTENCE\_TOKENS\_GO\_HERE>\newline

	JSON Output:\newline
\end{promptbox}
\caption{Example \textit{Basic} prompt for NoiseBench.}
\label{fig:prompt}
\end{figure}

\subsection{Models and Fine-Tuning}
We used \texttt{DistilBERT} and \texttt{XLM-RoBERTa (large)} for fine-tuning. We trained each model for 10 epochs, linear scheduler with 0.1 warmup, weight decay of 0.01. For NoiseBench and Wikigold, which include document boundaries, we added document context of 64 sentences. We ran each experiment for 3 random seeds. 

We tuned the learning rate and batch size for each dataset and model separately. We tried the following learning rate values: \texttt{[5.0e-6, 5.0e-5, 1.0e-5, 1.0e-6]} and train batch size values: \texttt{[8, 4, 16, 32]} and selected the best one based on the performance on the (noisy) development set\footnote{The best parameter sets can be found in our repository: \url{https://github.com/elenamer/error-type-aware-losses}}. 

\subsection{Baseline Losses}

For some related work losses, we performed hyperparameter tuning. For Focal loss we tried the following $\gamma$ values: \texttt{[0, 1, 2, 3]}. For GCE we tuned $q$ out of \texttt{[0.3, 0.5, 0.7, 0.9]}.

\subsection{Error-Type-Aware Losses}

For all error-type-aware losses, we use a threshold posterior cutoff of 0.3. With each loss, there is a warmup stage of 150 iterations, during which no masking is performed. Then, the confidence threshold is set to an initial value of 0.4 for $L_{\text{missing}}$ and 0.85 for $L_{\text{entity}}$. The threshold is updated three times during training, by fitting a BMM (10 iterations), after epochs 1, 3 and 6.

For $L_{\text{entity}}$, $L_{\text{all}}$, $L_{\text{FP}}$, $L_{\text{type}}$ and $L_{\text{missing,entity}}$ we implemented an entity-coverage guard, where the masking rule is applied only if a minimum non-O token predictions are made by the model (10\% of the observed non-O tokens in a given batch). This was necessary to prevent training collapse, which happens when the model has no positive (non-O) examples to learn from.

\section{LLM Error Propagation} 

We analyze the error propagation when fine-tuning on LLM annotated data. For this, we compare the labels obtained through direct prompting for the test set with the predictions from a model fine-tuned on the corresponding LLM-labeled training set. Table \ref{tab:overall-ce-only-llm-linked} shows, for each error type, the share of fine-tuned-model errors that occur at an LLM-error location (LLM-matched) versus errors made where the LLM test annotation was correct (FT-only). \textit{Missing} mentions and \textit{partial} matches are strongly propagated through fine-tuning, with around 80\% of these errors occurring at LLM error locations. In contrast, this number is 55\% for \textit{type} and 63\% for \textit{hallucinated} errors, which means they are often introduced independently by the fine-tuned model. Fine-tuning therefore largely inherits LLM missing and span-boundary annotation errors, while wrong types and hallucinations are less strongly associated with the LLM errors.
\input{tables/exp1_error_propagation_small_table}

\section{Results with \texttt{XLM-RoBERTa (large)}}
\label{appendix:xlm-roberta_results}
\input{tables/exp2_table_xlm-roberta}

\input{tables/exp2_comparison_to_other_losses_xlm-roberta}
All results in the main paper were from \texttt{DistilBERT}, with the exception of Figure \ref{fig:teaser}, which uses \texttt{XLM-RoBERTa (large)}. In this section we show the main results tables with an alternative model, \texttt{XLM-RoBERTa (large)}.

\section{Results on Extended Noisy Variants}
In this section, we show the results of our loss variants on the extended set of noisy variants (listed in Table \ref{tab:noise-levels-other-labels}) and on the dataset without noise. 
Table \ref{tab:distilbert-other-noises} shows the results with \texttt{DistilBERT} and Table \ref{tab:xlm-roberta-other-noises} shows the results with XLM-RoBERTa-Large.

\input{tables/exp2_table_other_noises_distilbert}
\input{tables/exp2_table_other_noises_xlm-roberta}


\end{document}

%% file: figures/introduction_figure.tex
\begin{figure}[tp!]
\vspace{-2mm}
\centering
\sffamily

\begin{subfigure}[b]{\linewidth}
\centering
\normalfont\small
\begin{tabular}{lccc}
\toprule
& Prompting & \multicolumn{2}{c}{Fine-Tuning w/}\\
Dataset & & LLM-labels & clean labels \\
\midrule
Ontonotes & 66.2 & 65.5 & \textbf{82.0} 
\\
NoiseBench & 83.5 & 84.4 & \textbf{94.8} 
\\
CDR & 74.2 & 71.2 & \textbf{85.4} 
\\
Wikigold & 73.4 & 70.3 & \textbf{79.7} 
\\
\bottomrule
\end{tabular}
\caption{\textit{Problem:} Noise from LLM-annotations transfers to fine-tuned models on such data.}\label{fig:teaser-ceiling}
\end{subfigure}

\vspace{6pt}

\begin{subfigure}[b]{\linewidth}
\centering
\begin{tikzpicture}[font=\sffamily]

\node[font=\small] at (0.0,6.10) {Albert};
\node[font=\small] at (1.1,6.10) {Einstein};
\node[font=\small] at (2.2,6.10) {won};
\node[font=\small] at (3.3,6.10) {the};
\node[font=\small] at (4.4,6.10) {Nobel};
\node[font=\small] at (5.5,6.10) {Prize};

\node[font=\scriptsize, anchor=east, text=gray] at (-0.7,5.6) {gold};
\node[font=\scriptsize] at (0.00,5.6) {B-PER};
\node[font=\scriptsize] at (1.10,5.6) {I-PER};
\node[font=\scriptsize] at (2.20,5.6) {O};
\node[font=\scriptsize] at (3.30,5.6) {O};
\node[font=\scriptsize] at (4.40,5.6) {B-MISC};
\node[font=\scriptsize] at (5.50,5.6) {I-MISC};

\node[font=\scriptsize, anchor=east, text=gray] at (-0.7,5.25) {pred};
\node[font=\scriptsize, text=teal!60!black] at (0.00,5.25) {B-PER};
\node[font=\scriptsize, text=teal!60!black] at (1.10,5.25) {I-PER};
\node[font=\scriptsize, text=teal!60!black] at (2.20,5.25) {O};
\node[font=\scriptsize, text=teal!60!black] at (3.30,5.25) {O};
\node[font=\scriptsize, text=teal!60!black] at (4.40,5.25) {B-MISC};
\node[font=\scriptsize, text=teal!60!black] at (5.50,5.25) {I-MISC};

\node[font=\scriptsize, anchor=east, text=gray] at (-0.7,4.75) {loss};
\foreach \x in {0.00,1.10,2.20,3.30,4.40,5.50}{
  \fill[gray!45]       (\x-0.20,4.55) rectangle (\x-0.02,4.95);
  \fill[blue!70!black] (\x+0.02,4.55) rectangle (\x+0.20,4.95);
}

\draw[-{Latex[length=1.6mm]}, gray!65, line width=0.7pt] (2.25,4.3) -- (2.25,3.7);
\node[font=\scriptsize, text=gray, align=center] at (3.55,4.0) {after some training};

\node[font=\small] at (0.00,3.40) {Curie};
\node[font=\small] at (0.90,3.40) {won};
\node[font=\small] at (1.80,3.40) {the};
\node[font=\small] at (2.70,3.40) {Nobel};
\node[font=\small] at (3.60,3.40) {Prize};
\node[font=\small] at (4.50,3.40) {in};
\node[font=\small] at (5.40,3.40) {Physics};

\node[font=\scriptsize, anchor=east, text=gray] at (-0.7,2.95) {gold};
\node[font=\scriptsize, text=orange!80!black] at (0.00,2.95) {O};
\node[font=\scriptsize]                       at (0.90,2.95) {O};
\node[font=\scriptsize]                       at (1.80,2.95) {O};
\node[font=\scriptsize, text=orange!80!black] at (2.70,2.95) {B-ORG};
\node[font=\scriptsize, text=orange!80!black] at (3.60,2.95) {I-ORG};
\node[font=\scriptsize]                       at (4.50,2.95) {O};
\node[font=\scriptsize, text=orange!80!black] at (5.40,2.95) {B-ORG};

\node[font=\scriptsize, anchor=east, text=gray] at (-0.7,2.60) {pred};
\node[font=\scriptsize, text=teal!60!black] at (0.00,2.60) {B-PER};
\node[font=\scriptsize, text=teal!60!black] at (0.90,2.60) {O};
\node[font=\scriptsize, text=teal!60!black] at (1.80,2.60) {O};
\node[font=\scriptsize, text=teal!60!black] at (2.70,2.60) {B-MISC};
\node[font=\scriptsize, text=teal!60!black] at (3.60,2.60) {I-MISC};
\node[font=\scriptsize, text=teal!60!black] at (4.50,2.60) {O};
\node[font=\scriptsize, text=teal!60!black] at (5.40,2.60) {O};

\node[font=\scriptsize, anchor=east, text=gray] at (-0.7,2.00) {loss};
\fill[gray!45]       (-0.20,1.85) rectangle (-0.02,2.25);
\fill[blue!70!black] ( 0.02,1.85) rectangle ( 0.20,1.89);
\fill[gray!45]       ( 0.70,1.85) rectangle ( 0.88,2.25);
\fill[blue!70!black] ( 0.92,1.85) rectangle ( 1.10,2.25);
\fill[gray!45]       ( 1.60,1.85) rectangle ( 1.78,2.25);
\fill[blue!70!black] ( 1.82,1.85) rectangle ( 2.00,2.25);
\fill[gray!45]       ( 2.50,1.85) rectangle ( 2.68,2.25);
\fill[blue!70!black] ( 2.72,1.85) rectangle ( 2.90,1.93);
\fill[gray!45]       ( 3.40,1.85) rectangle ( 3.58,2.25);
\fill[blue!70!black] ( 3.62,1.85) rectangle ( 3.80,1.95);
\fill[gray!45]       ( 4.30,1.85) rectangle ( 4.48,2.25);
\fill[blue!70!black] ( 4.52,1.85) rectangle ( 4.70,2.25);
\fill[gray!45]       ( 5.20,1.85) rectangle ( 5.38,2.25);
\fill[blue!70!black] ( 5.42,1.85) rectangle ( 5.60,1.97);

\node[font=\scriptsize, anchor=west] at (1.0,1.50)
  {\textcolor{gray!45}{\rule{5pt}{5pt}}~cross entropy\quad
   \textcolor{blue!70!black}{\rule{5pt}{5pt}}~ours};

\end{tikzpicture}
\caption{\textit{Approach:} Our proposed loss ignore erroneous annotations after some training.}\label{fig:teaser-masking}
\end{subfigure}

\vspace{6pt}

\begin{subfigure}[b]{\linewidth}
\centering
\begin{tikzpicture}[font=\sffamily]
\begin{axis}[
  width=6.0cm, height=2.5cm, scale only axis,
  font=\sffamily,
  xlabel={label noise}, ylabel={F1},
  xlabel style={font=\small, yshift=2pt},
  ylabel style={font=\small, yshift=-3pt},
  tick label style={font=\small},
  xmin=15, xmax=35, ymin=60, ymax=75,
  xtick={15,20,25,30,35}, xticklabels={15\%,20\%,25\%,30\%,35\%},
  ytick={65,70,75}, yticklabels={65,70,75}, ytick align=outside,
  axis lines=left,
  axis line style={-{Latex[length=1.4mm]}, gray!70, line width=0.5pt},
  legend style={font=\scriptsize, at={(0.03,0.05)}, anchor=south west,
                draw=none, fill=none, row sep=-2pt},
  legend cell align=left,
]
\addplot[blue!60!black, line width=1.2pt, mark=*, mark size=1.8pt]
  coordinates { (33.4,67.8) (28.3,65.8) (24.1,67.9) (20.4,72.9) }; 
\addlegendentry{ours}
\addplot[gray!55, line width=1.2pt, mark=*, mark size=1.8pt]
  coordinates {(33.4,64.7) (28.3,65.0) (24.1,66.5) (20.4,70.3) }; 
\addlegendentry{cross entropy}


\end{axis}
\end{tikzpicture}
\caption{\textit{Results:} Across different noise-level, we observe that our loss outperforms comparable loss functions. Plot shows the results with $L_{\text{missing}}$ on Wikigold.}\label{fig:teaser-noise}
\end{subfigure}

\caption{We introduce a error-type aware loss formulation for the task of named entity recognition as we find that noisy LLM-annotated datasets directly impact downstream performance. Our loss function helps to overcome this issue and improve over comparable loss functions across a range of noise levels.}
\vspace{-1mm}
\label{fig:teaser}
\end{figure}
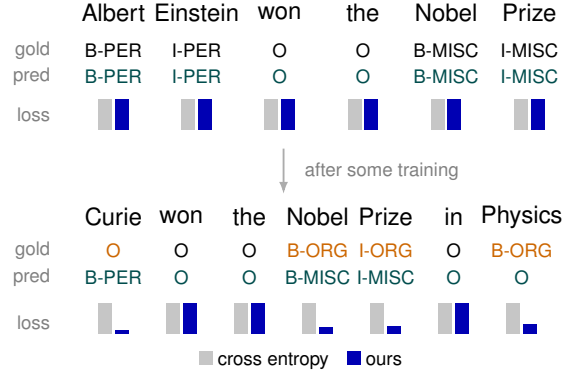
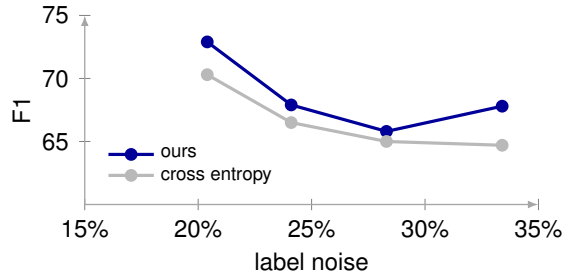

%% file: tables/exp1_datasets_overview_llm.tex
\definecolor{type}{HTML}{E8F5E9}
\definecolor{FN}{HTML}{FFF4CC}
\definecolor{FP}{HTML}{FCE4EC}
\begin{table}[htbp!]
\centering
\small
\setlength{\tabcolsep}{4pt}
\begin{tabular}{lccccc}
\toprule
& \textit{F1}& \multicolumn{4}{c}{\textit{\%Errors}}\\
Dataset &  & \cellcolor{FN}{\textit{FN}} & \cellcolor{FP}{\textit{FP}} & \cellcolor{type}{\textit{Type}} & \textit{Partial} \\
\midrule
\multicolumn{6}{l}{\textit{OntoNotes}} \\
GPT-OSS EvoPrompt & 70.8 & \cellcolor{FN}{31.1} & 29.4 & 9.3 & 30.2 \\
GPT-OSS Basic & 67.0 & 31.8 & \cellcolor{FP}{35.7} & 8.6 & 23.9 \\
Qwen DiZiNER & 60.7 & \cellcolor{FN}{58.3} & 16.3 & 5.0 & 20.4 \\
\midrule
\multicolumn{6}{l}{\textit{NoiseBench (CoNLL)}} \\
GPT-OSS Schema & 85.0 & 27.2 & 16.1 & \cellcolor{type}{28.5} & 28.2 \\
GPT-OSS EvoPrompt & 80.5 & 20.7 & 18.9 & \cellcolor{type}{50.2} & 10.2 \\
GPT-OSS Basic & 75.1 & 17.0 & \cellcolor{FP}{43.1} & 31.9 & 8.0 \\
\midrule
\multicolumn{6}{l}{\textit{BC5CDR}} \\
GPT-OSS Schema & 74.1 & \cellcolor{FN}{63.0} & 18.5 & 0.3 & 18.2 \\
GPT-OSS DiZiNER & 73.0 & \cellcolor{FN}{53.5} & 12.3 & 0.3 & 33.9 \\
Qwen DiZiNER & 69.6 & \cellcolor{FN}{49.3} & 33.4 & 0.6 & 16.7 \\
\midrule
\multicolumn{6}{l}{\textit{Wikigold}} \\
GPT-OSS Schema & 79.6 & 26.7 & \cellcolor{FP}{32.0} & 24.7 & 16.6 \\
GPT-OSS EvoPrompt & 75.9 & 13.5 & \cellcolor{FP}{64.3} & 11.8 & 10.4 \\
Qwen DiZiNER & 72.7 & 34.0 & \cellcolor{FP}{46.2} & 11.9 & 7.9 \\
\bottomrule
\end{tabular}
\caption{Overview of LLM label noise of different label variants in terms of F1 and error shares. The error types are: \textit{FN} (missing mentions), \textit{FP} (hallucinated false positive mentions), \textit{Type} - (correct boundary, but wrong type) and \textit{Partial} (correct type, but partially wrong boundary). The error type representing the largest share of errors is color-highlighted.}
\label{tab:noise-levels-llm-labels}
\end{table}

%% file: tables/exp2_table_distilbert.tex
\definecolor{type}{HTML}{E8F5E9}
\definecolor{FN}{HTML}{FFF4CC}
\definecolor{FP}{HTML}{FCE4EC}

\newcommand{\splitcell}[1]{%
  \tikz[baseline=(M.base)] \node[
    inner sep=3pt, font=\small,
    path picture={
      \fill[FP] (path picture bounding box.north west)
        rectangle ($(path picture bounding box.north east)!0.5!(path picture bounding box.south east)$);
      \fill[type] ($(path picture bounding box.north west)!0.5!(path picture bounding box.south west)$)
        rectangle (path picture bounding box.south east);
    }
  ] (M) {#1};
}
  
\begin{table*}[t]
\centering
\small
\setlength{\tabcolsep}{3.5pt}
\begin{tabular}{llllllll}
\toprule
Label set & CE & $L_{\text{all}}$ & \cellcolor{FN}{$L_{\text{missing}}$} & \splitcell{$L_{\text{entity}~~~~~}$} & \cellcolor{FP}{$L_{\text{FP}}$} & \cellcolor{type}{$L_{\text{type}}$} & $L_{\text{missing,entity}}$ \\
\midrule
\multicolumn{8}{l}{\textit{BC5CDR}} \\
\cellcolor{FN}{GPT-OSS Schema} & 70.1 $\pm$ 0.3 & 61.3 $\pm$ 6.3 & \cellcolor{gray!15}70.6 $\pm$ 0.5 & 68.2 $\pm$ 0.3 & 67.9 $\pm$ 0.2 & 69.6 $\pm$ 0.2 & 65.9 $\pm$ 1.2 \\
\cellcolor{FN}{GPT-OSS DiZiNER} & 70.1 $\pm$ 0.3 & 68.8 $\pm$ 2.2 & \cellcolor{gray!15}71.3 $\pm$ 0.3 & 70.5 $\pm$ 0.3 & 69.7 $\pm$ 0.1 & 70.2 $\pm$ 0.3 & 68.7 $\pm$ 1.2 \\
\cellcolor{FN}{Qwen DiZiNER} & 68.7 $\pm$ 0.2 & 63.1 $\pm$ 4.4 & \cellcolor{gray!15}70.4 $\pm$ 0.3 & 67.8 $\pm$ 0.4 & 67.5 $\pm$ 0.2 & 68.0 $\pm$ 0.6 & 67.0 $\pm$ 0.2 \\
Average & 69.6 $\pm$ 0.7 & 64.4 $\pm$ 3.2 & \cellcolor{gray!15}70.7 $\pm$ 0.4 & 68.8 $\pm$ 1.2 & 68.4 $\pm$ 0.9 & 69.3 $\pm$ 0.9 & 67.2 $\pm$ 1.2 \\
\midrule
\multicolumn{8}{l}{\textit{NoiseBench}} \\
\cellcolor{type}{GPT-OSS Schema} & 80.2 $\pm$ 0.3 & 80.0 $\pm$ 0.9 & 80.0 $\pm$ 0.7 & \cellcolor{gray!15}81.8 $\pm$ 0.6 & 79.8 $\pm$ 0.8 & 81.7 $\pm$ 0.1 & 79.6 $\pm$ 1.7 \\
\cellcolor{type}{GPT-OSS EvoPrompt} & 72.0 $\pm$ 0.2 & 71.5 $\pm$ 1.1 & 72.3 $\pm$ 0.2 & \cellcolor{gray!15}72.8 $\pm$ 0.7 & 72.2 $\pm$ 0.1 & 72.0 $\pm$ 0.4 & 70.8 $\pm$ 0.8 \\
\cellcolor{FP}{GPT-OSS Basic} & 68.9 $\pm$ 0.6 & 69.3 $\pm$ 1.4 & 68.3 $\pm$ 0.3 & \cellcolor{gray!15}72.5 $\pm$ 0.5 & 72.2 $\pm$ 0.2 & 68.2 $\pm$ 0.9 & 69.0 $\pm$ 2.6 \\
Average & 73.7 $\pm$ 4.8 & 73.6 $\pm$ 4.6 & 73.5 $\pm$ 4.9 & \cellcolor{gray!15}75.7 $\pm$ 4.3 & 74.7 $\pm$ 3.6 & 73.9 $\pm$ 5.7 & 73.1 $\pm$ 4.6 \\
\midrule
\multicolumn{8}{l}{\textit{OntoNotes}} \\
\cellcolor{FN}{GPT-OSS EvoPrompt} & 61.5 $\pm$ 0.1 & 59.7 $\pm$ 2.0 & \cellcolor{gray!15}62.1 $\pm$ 0.3 & 59.9 $\pm$ 0.3 & 61.4 $\pm$ 0.3 & 61.2 $\pm$ 0.1 & 58.8 $\pm$ 0.3 \\
\cellcolor{FP}{GPT-OSS Basic} & 57.6 $\pm$ 0.3 & 55.4 $\pm$ 2.2 & \cellcolor{gray!15}58.0 $\pm$ 0.7 & 54.6 $\pm$ 1.1 & 57.2 $\pm$ 0.5 & 57.5 $\pm$ 0.4 & 54.9 $\pm$ 1.2 \\
\cellcolor{FN}{Qwen DiZiNER} & 57.0 $\pm$ 0.4 & 55.5 $\pm$ 1.3 & \cellcolor{gray!15}58.3 $\pm$ 0.3 & 55.3 $\pm$ 0.6 & 56.1 $\pm$ 0.4 & 57.1 $\pm$ 0.2 & 56.4 $\pm$ 0.6 \\
Average & 58.7 $\pm$ 2.0 & 56.9 $\pm$ 2.0 & \cellcolor{gray!15}59.5 $\pm$ 1.9 & 56.6 $\pm$ 2.3 & 58.2 $\pm$ 2.2 & 58.6 $\pm$ 1.8 & 56.7 $\pm$ 1.6 \\
\midrule
\multicolumn{8}{l}{\textit{Wikigold}} \\
\cellcolor{FP}{GPT-OSS Schema} & 66.8 $\pm$ 0.2 & \cellcolor{gray!15}70.3 $\pm$ 2.1 & 70.2 $\pm$ 1.0 & 68.6 $\pm$ 1.2 & 70.0 $\pm$ 0.9 & 67.8 $\pm$ 1.4 & 67.0 $\pm$ 1.3 \\
\cellcolor{FP}{GPT-OSS EvoPrompt} & 60.2 $\pm$ 0.5 & 60.4 $\pm$ 2.5 & 59.3 $\pm$ 0.8 & 63.3 $\pm$ 1.7 & \cellcolor{gray!15}64.8 $\pm$ 1.8 & 61.6 $\pm$ 0.5 & 60.3 $\pm$ 0.8 \\
\cellcolor{FP}{Qwen DiZiNER} & \cellcolor{gray!15}59.8 $\pm$ 0.2 & 56.4 $\pm$ 1.3 & 59.7 $\pm$ 0.9 & 57.7 $\pm$ 1.2 & 55.6 $\pm$ 1.1 & 58.8 $\pm$ 1.0 & 59.0 $\pm$ 0.6 \\
Average & 62.2 $\pm$ 3.2 & 62.4 $\pm$ 5.8 & 63.1 $\pm$ 5.1 & 63.2 $\pm$ 4.5 & \cellcolor{gray!15}63.5 $\pm$ 6.0 & 62.7 $\pm$ 3.8 & 62.1 $\pm$ 3.5 \\
\bottomrule
\end{tabular}
\caption{Test F1s (entity-level) of error-type-aware losses, compared with cross-entropy (CE) and global masking ($L_{\text{all}}$), using \texttt{DistilBERT}. The color-highlighting is consistent with Table \ref{tab:noise-levels-llm-labels} and refers to the most common error type in each label variant (rows) and the error types targeted by each loss (columns).}
\label{tab:distilbert-base-uncased-test-f1-wide}
\end{table*}

%% file: figures/lineplot_noisebench_F1s.tex
\begin{figure*}[htbp!]
    \centering
    \includegraphics[width=0.8\textwidth]{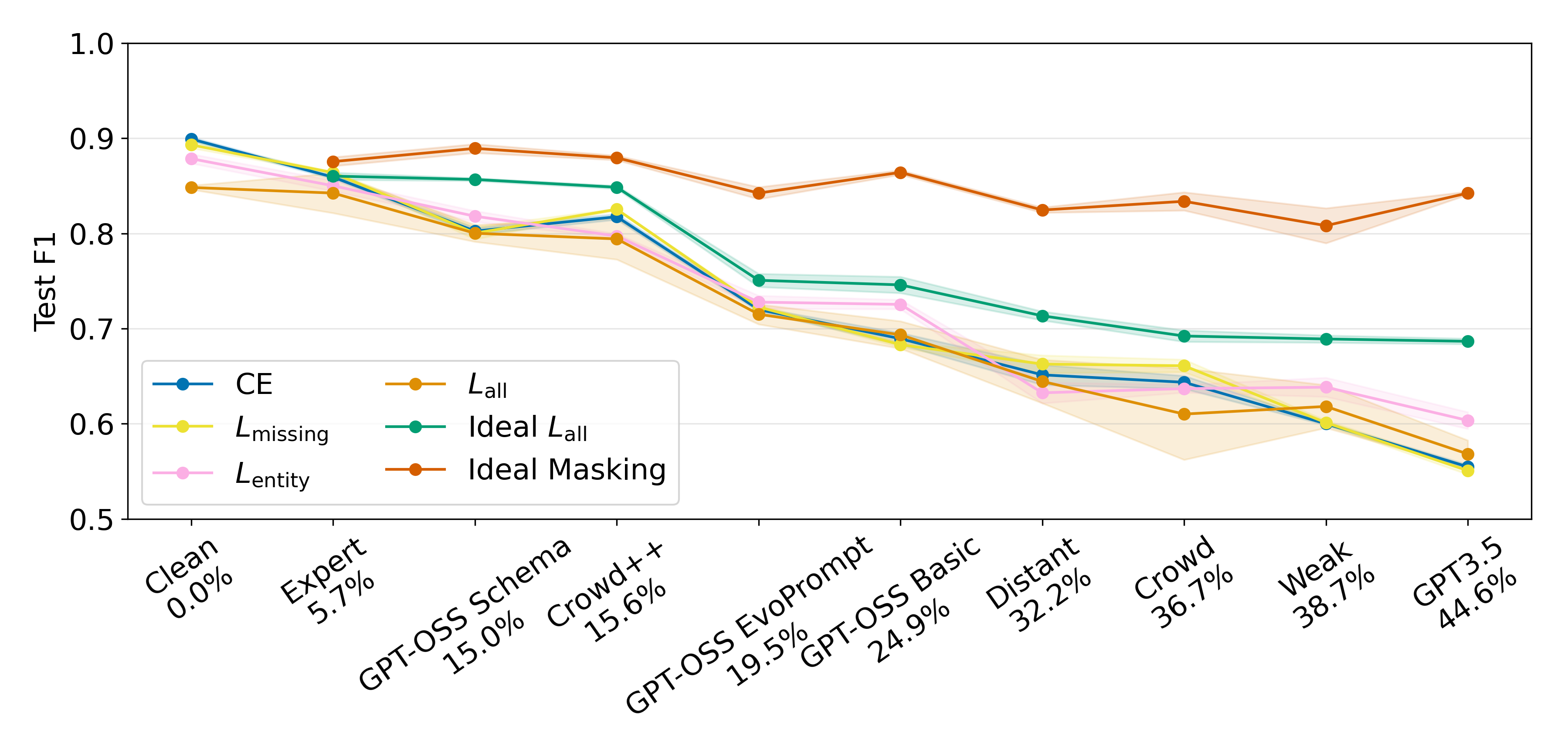}
    \caption{Test F1s across noise levels, of error-type-aware losses, CE and oracle masking. Shows all 10 NoiseBench noisy variants, using \texttt{DistilBERT}. The noise share is calculated as $1 - F1_{\text{annotation}}$. }
    \label{fig:F1s_lineplot}
\end{figure*}

%% file: tables/exp2_comparison_to_other_losses_distilbert.tex
\begin{table}[t]
\centering
\small
\setlength{\tabcolsep}{2.65pt}
\centering
\begin{tabular}{lcccc}
\toprule
Loss & BC5CDR & NoiseBench & OntoNotes & Wikigold \\
\midrule
\multicolumn{3}{l}{\textit{Baseline losses}} \\
CE & 69.6 $\pm$ 0.7 & 73.7 $\pm$ 4.8 & 58.7 $\pm$ 2.0 & 62.2 $\pm$ 3.2 \\
$\text{BMM}_{\text{b}}$ & 69.4 $\pm$ 0.9 & 73.7 $\pm$ 4.8 & 58.8 $\pm$ 2.2 & 62.2 $\pm$ 4.2 \\
$\text{NLL}_{\text{c}}$ & 69.4 $\pm$ 0.9 & 73.9 $\pm$ 5.0 & 58.8 $\pm$ 2.1 & 62.3 $\pm$ 3.5 \\
GCE & 69.3 $\pm$ 0.9 & 73.3 $\pm$ 5.1 & 58.7 $\pm$ 2.1 & 62.5 $\pm$ 3.5 \\
Focal & 69.6 $\pm$ 0.9 & 73.7 $\pm$ 4.9 & 58.8 $\pm$ 2.2 & 63.1 $\pm$ 4.3 \\
$L_{\text{all}}$ & 64.4 $\pm$ 3.2 & 73.6 $\pm$ 4.6 & 56.9 $\pm$ 2.0 & 62.4 $\pm$ 5.8 \\
\midrule
\multicolumn{3}{l}{\textit{Error-type-aware losses}} \\
$L_{\text{missing}}$ & \cellcolor{gray!15}70.7 $\pm$ 0.4 & 73.5 $\pm$ 4.9 & \cellcolor{gray!15}59.5 $\pm$ 1.9 & 63.1 $\pm$ 5.1 \\
$L_{\text{entity}}$ & 68.8 $\pm$ 1.2 & \cellcolor{gray!15}75.7 $\pm$ 4.3 & 56.6 $\pm$ 2.3 & 63.2 $\pm$ 4.5 \\
$L_{\text{FP}}$ & 68.4 $\pm$ 0.9 & 74.7 $\pm$ 3.6 & 58.2 $\pm$ 2.2 & \cellcolor{gray!15}63.5 $\pm$ 6.0 \\
$L_{\text{type}}$ & 69.3 $\pm$ 0.9 & 73.9 $\pm$ 5.7 & 58.6 $\pm$ 1.8 & 62.7 $\pm$ 3.8 \\
$L_{\text{miss., ent.}}$ & 67.2 $\pm$ 1.2 & 73.1 $\pm$ 4.6 & 56.7 $\pm$ 1.6 & 62.1 $\pm$ 3.5 \\
\bottomrule
\end{tabular}
\caption{Comparison of error-aware losses with other related noise-robust losses w.r.t. test F1, using \texttt{DistilBERT}. }
\label{tab:distilbert-base-uncased-test-f1-dataset-averages-wide}
\end{table}

%% file: figures/thresholds_pu_loss.tex
\begin{figure*}[htp]
    \centering
    \begin{subfigure}[t]{0.49\textwidth}
        \centering
        \includegraphics[width=\textwidth]{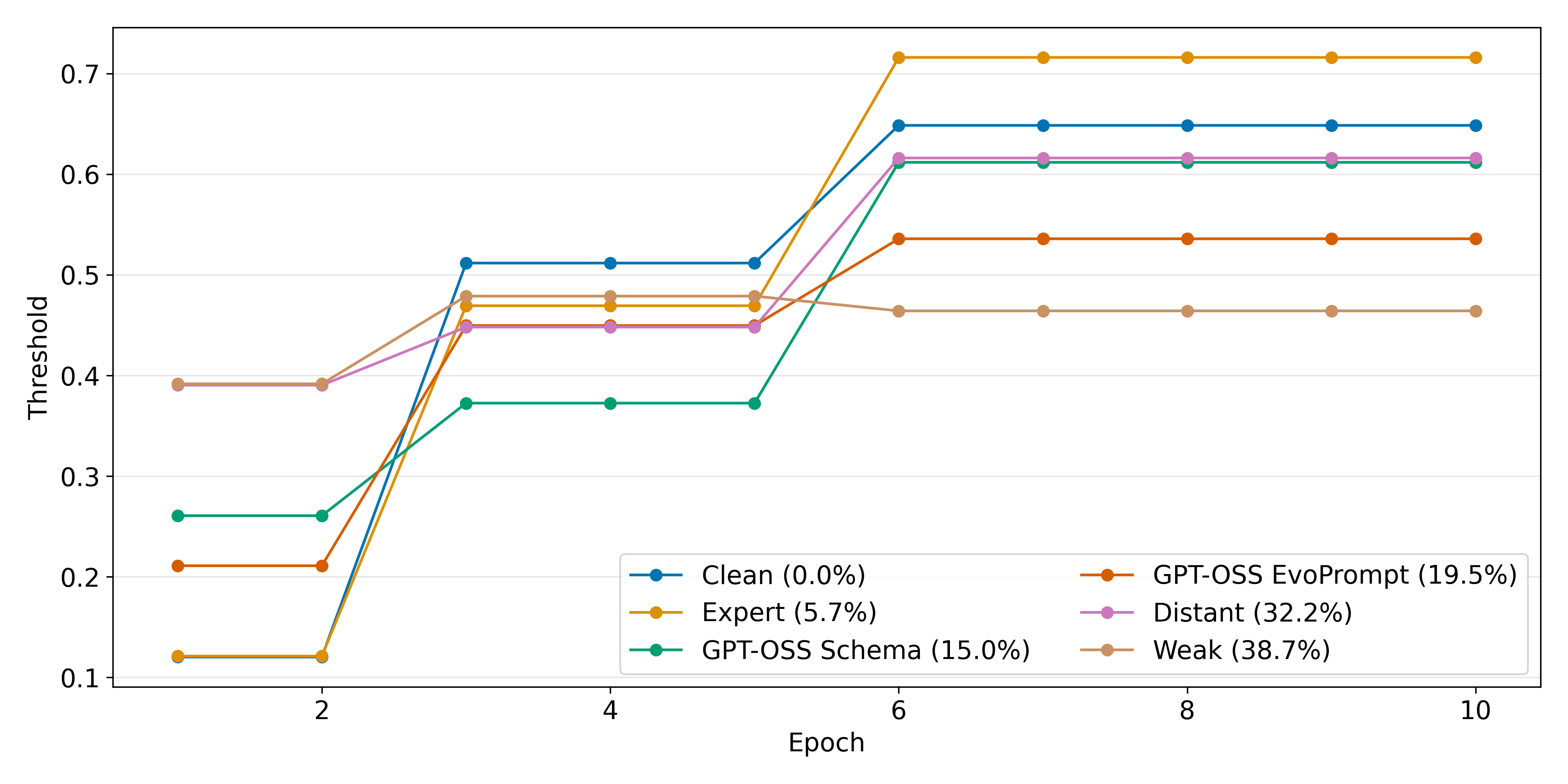}
        \caption{Confidence threshold.}
    \label{fig:pu-loss-thresholds}
    \end{subfigure}%
    ~ 
    \begin{subfigure}[t]{0.49\textwidth}
        \centering
        \includegraphics[width=\textwidth]{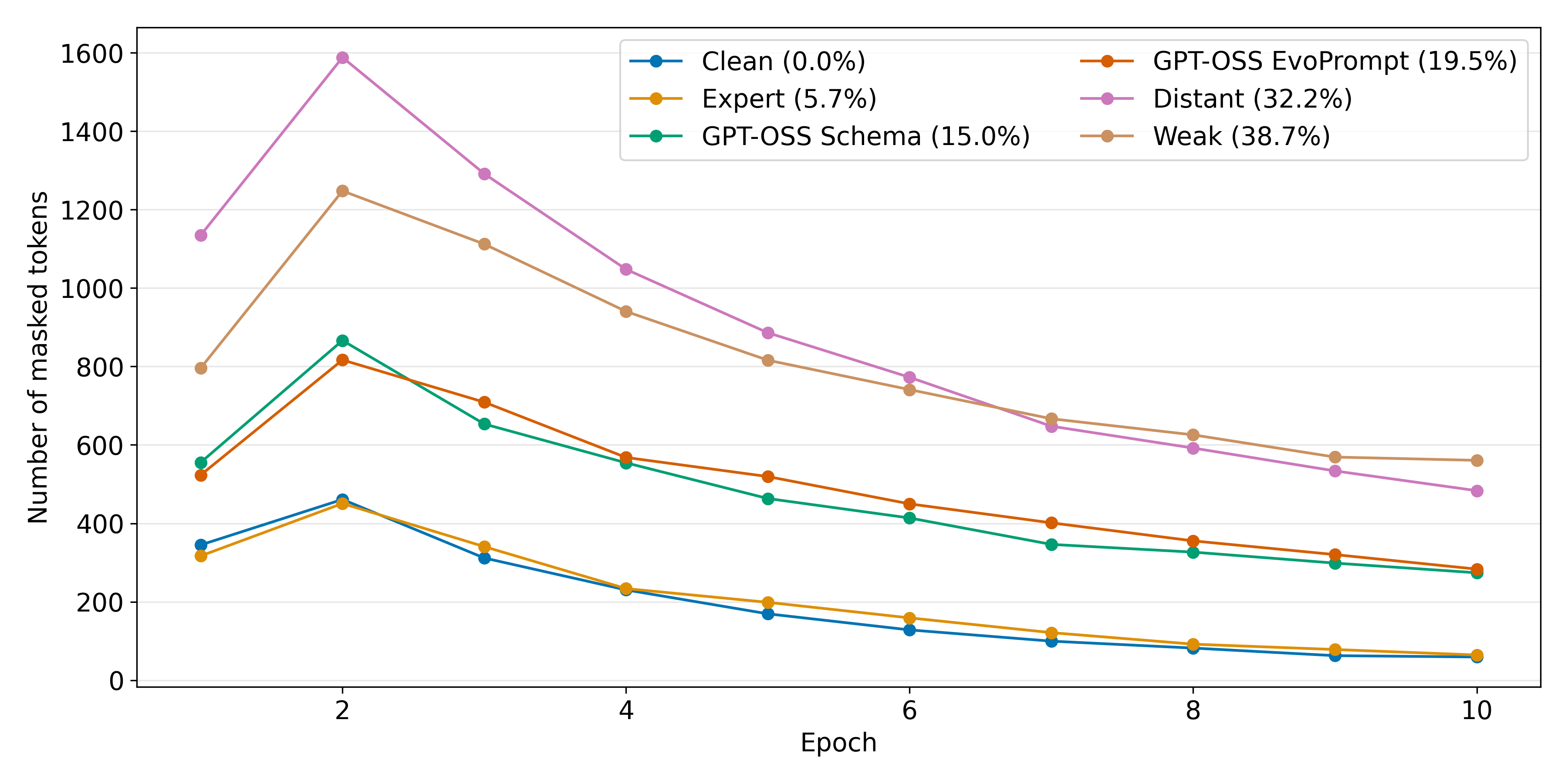}
        \caption{Number of masked tokens.}
    \label{fig:pu-masked-tokens}

    \end{subfigure}
    \caption{Confidence threshold and number of masked tokens with $L_{\text{missing}}$ loss, using \texttt{DistilBERT}. Shows selected NoiseBench noisy variants.}
\end{figure*}

%% file: figures/combined_losses_plots.tex
\begin{figure*}[htp]
    \centering
    \begin{subfigure}[t]{0.32\textwidth}
        \centering
        \includegraphics[width=\textwidth]{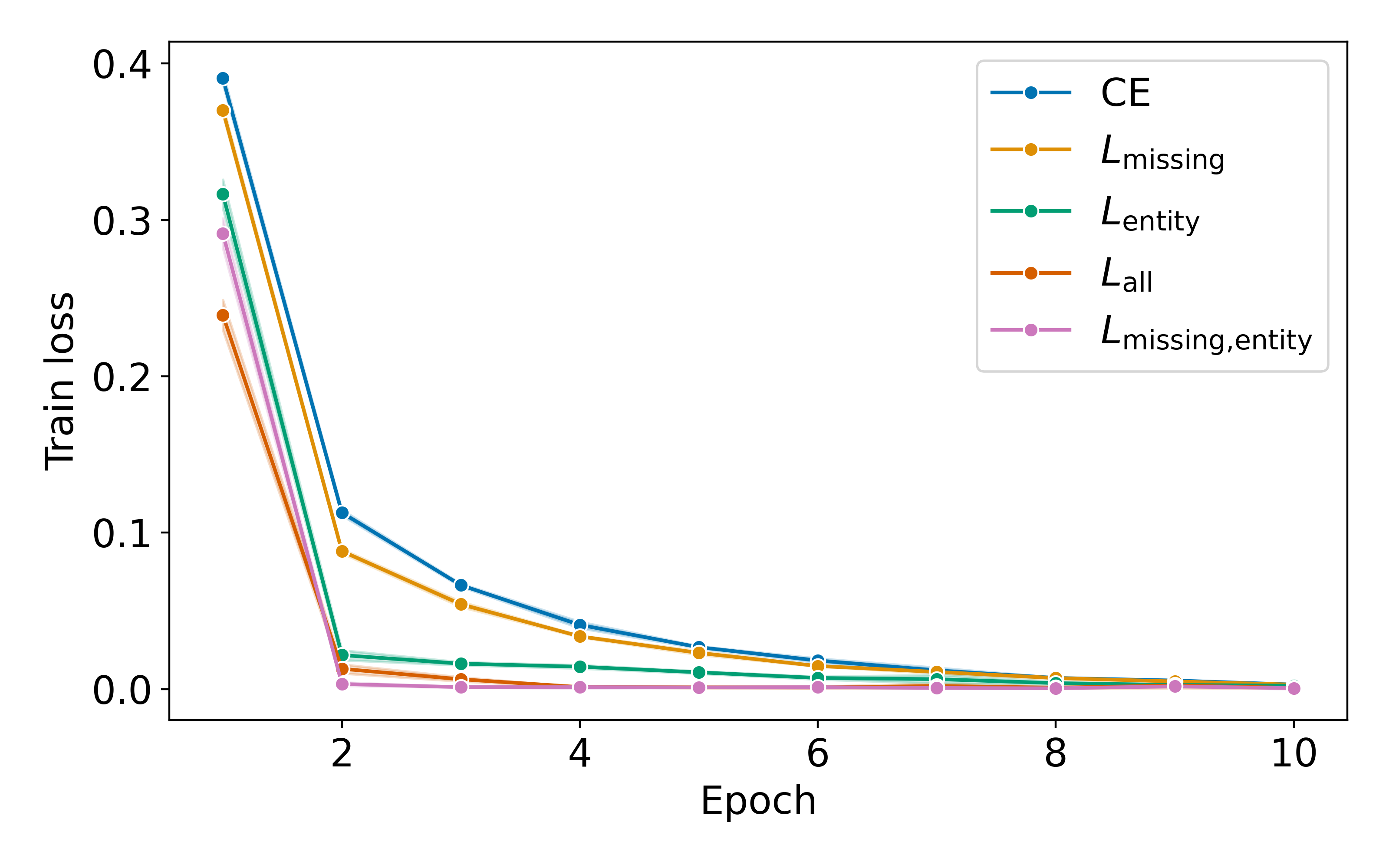}
        \caption{Training loss.}
    \label{fig:loss}
    \end{subfigure}%
    ~ 
    \begin{subfigure}[t]{0.32\textwidth}
        \centering
        \includegraphics[width=\textwidth]{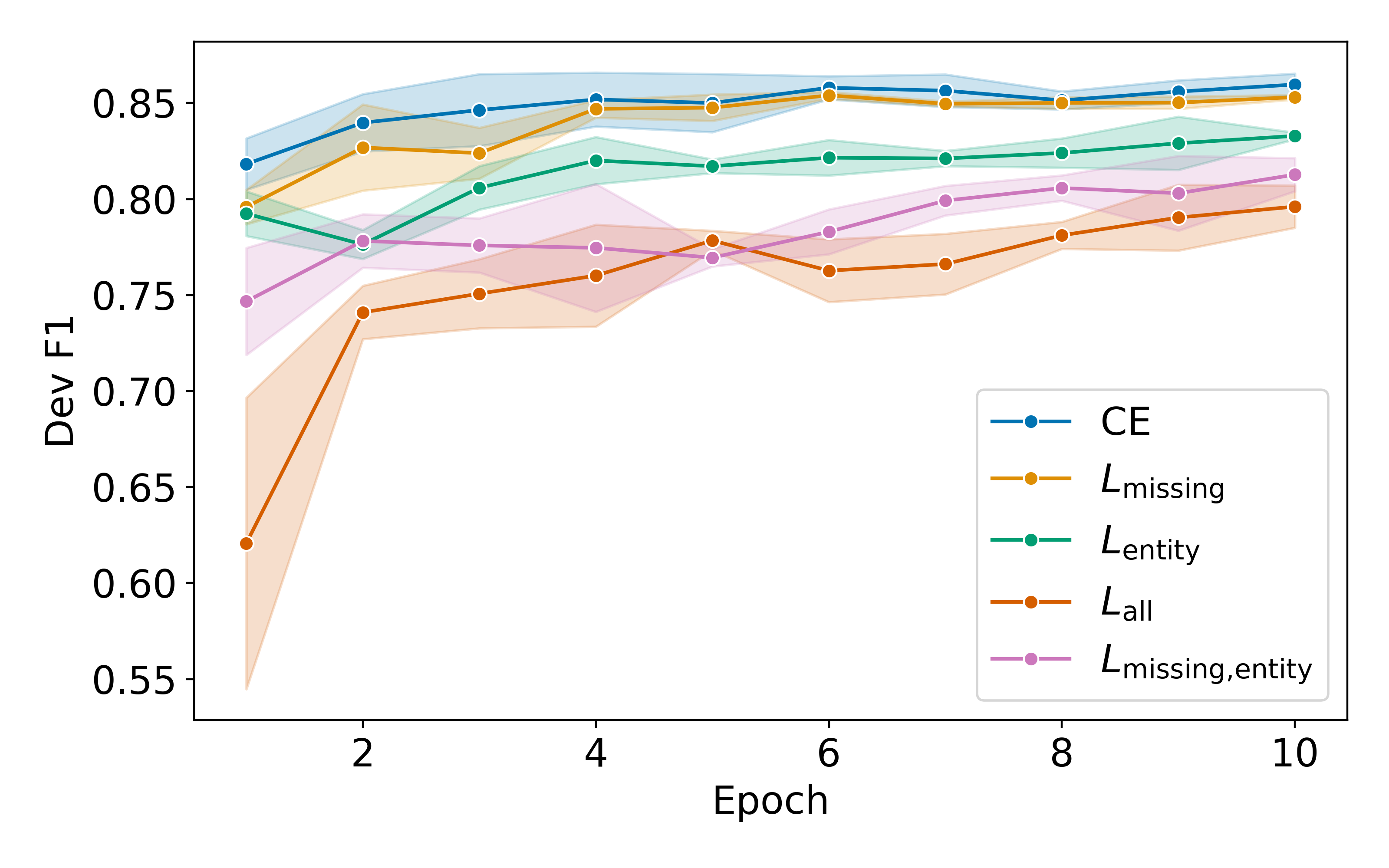}
        \caption{Development F1.}
    \label{fig:dev-f1}

    \end{subfigure}
    ~    
    \begin{subfigure}[t]{0.32\textwidth}
        \centering
        \includegraphics[width=\textwidth]{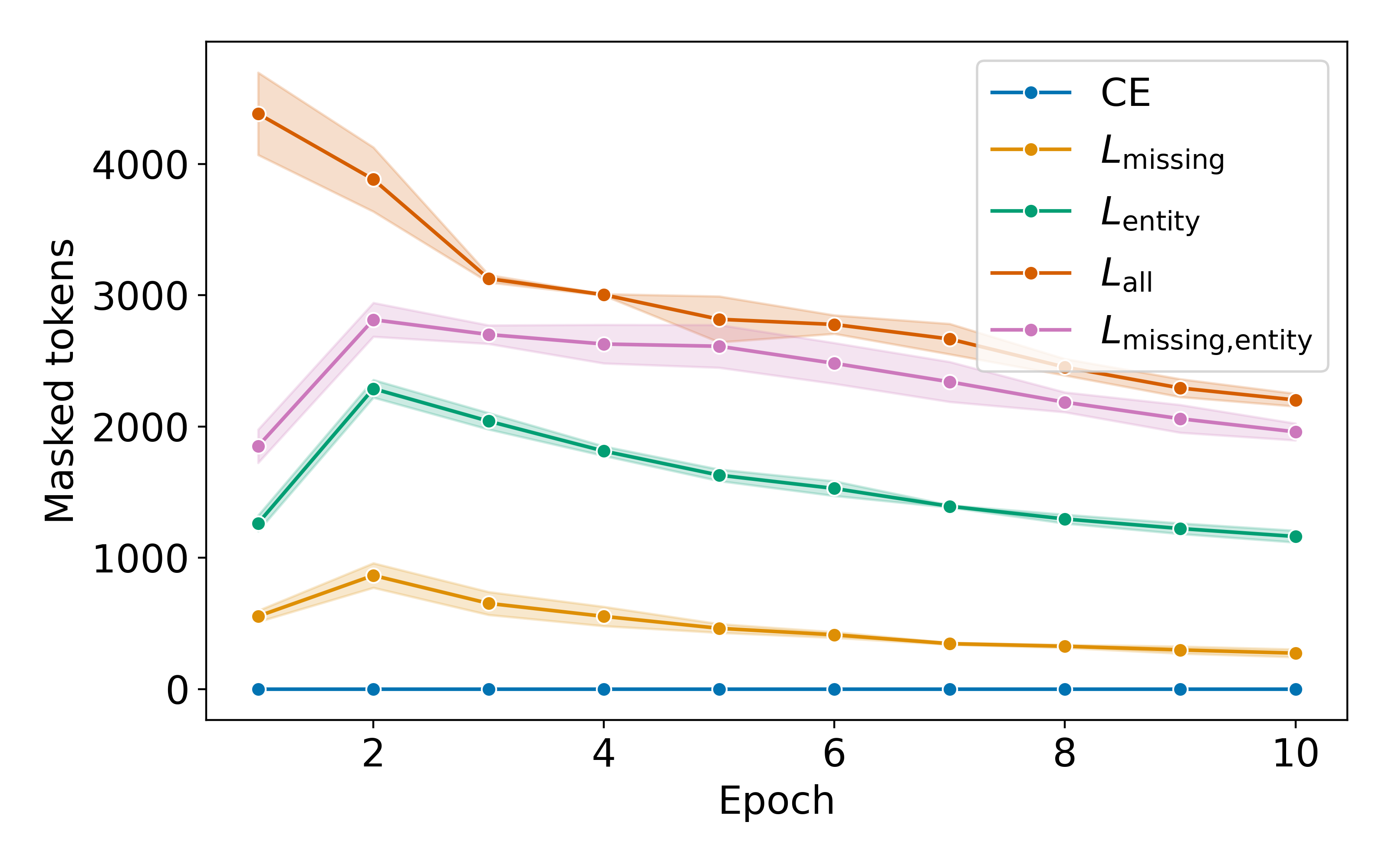}
        \caption{Number of masked tokens.}
    \label{fig:masked-tokens}

    \end{subfigure}
    \caption{Training dynamics of different loss variants, when fine-tuning on NoiseBench's \texttt{GPT-OSS Schema} variant. }
    \label{fig:training-dynamics}

\end{figure*}

%% file: tables/appendix_dataset_sizes.tex
\begin{table}[htp]
\centering
\small
\begin{tabular}{lrrr}
\toprule
& \multicolumn{2}{c}{\# Sentences} & \#Entity \\
Dataset & Train & Test & Types  \\
\midrule
OntoNotes & 15036 & 9479 & 18 \\
NoiseBench & 4879 & 3426 & 4 \\
BC5CDR & 4576 & 4784 & 2 \\
Wikigold & 1330 & 120 & 4 \\
\bottomrule
\end{tabular}
\caption{Dataset sizes and numbers of entity classes.}
\label{tab:dataset-sizes}
\end{table}

%% file: tables/exp1_datasets_overview_other.tex
\begin{table}[htp]
\centering
\small
\setlength{\tabcolsep}{4pt}
\begin{tabular}{lrrrrr}
\toprule
& \textit{F1}& \multicolumn{4}{c}{\textit{\%Errors}}\\
Dataset &  & {\textit{FN}} & {\textit{FP}} & {\textit{Type}} & \textit{Partial} \\
\midrule
\multicolumn{6}{l}{\textit{OntoNotes}} \\
Distant & 75.9 & 55.1 & 17.6 & 9.9 & 17.4 \\
\midrule
\multicolumn{6}{l}{\textit{NoiseBench}} \\
Expert & 94.3 & 9.7 & 3.6 & 73.4 & 13.3 \\
Crowd++ & 84.4 & 58.6 & 9.1 & 16.7 & 15.6 \\
Distant & 67.8 & 65.5 & 10.5 & 12.9 & 11.1 \\
Crowd & 63.3 & 61.8 & 10.5 & 15.1 & 12.6 \\
Weak & 61.3 & 17.4 & 36.5 & 33.7 & 12.3 \\
GPT3.5 & 55.4 & 23.5 & 46.8 & 25.8 & 3.9 \\
\midrule
\multicolumn{6}{l}{\textit{BC5CDR}} \\
Distant & 79.0 & 90.2 & 6.9 & 0.1 & 2.8 \\
\midrule
\multicolumn{6}{l}{\textit{Wikigold}} \\
Distant & 66.6 & 35.3 & 37.3 & 10.7 & 16.7 \\
\bottomrule
\end{tabular}
\caption{Overview of the noisy training variants, from extended noise sources.}
\label{tab:noise-levels-other-labels}
\end{table}

%% file: tables/exp1_error_propagation_small_table.tex
\begin{table}[t]
\centering
\small

\begin{tabular}{lcc}
\toprule
Error type & FT-only \% & LLM-matched \% \\
\midrule
Missing (FN) & 21.1 & 78.9 \\
Hallucinated (FP) & 36.6 & 63.4 \\
Type & 44.9 & 55.1 \\
Partial & 18.9 & 81.1 \\
\bottomrule
\end{tabular}
\caption{Overall FT-only and LLM-matched shares by error type, for a \texttt{XLM-RoBERTa (large)}.}
\label{tab:overall-ce-only-llm-linked}
\end{table}

%% file: tables/exp2_table_xlm-roberta.tex
\definecolor{type}{HTML}{E8F5E9}
\definecolor{FN}{HTML}{FFF4CC}
\definecolor{FP}{HTML}{FCE4EC}
\begin{table*}[t]
\centering
\small
\setlength{\tabcolsep}{3pt}

\begin{tabular}{llllllll}
\toprule
Label variant & CE & $L_{\text{all}}$ & $L_{\text{missing}}$ & $L_{\text{entity}}$ & $L_{\text{FP}}$ & $L_{\text{type}}$ & $L_{\text{missing,entity}}$ \\
\midrule
\multicolumn{8}{l}{\textit{BC5CDR}} \\
GPT-OSS Schema & 71.2 $\pm$ 0.3 & 67.1 $\pm$ 1.2 & \cellcolor{gray!15}73.0 $\pm$ 0.1 & 67.9 $\pm$ 0.3 & 66.9 $\pm$ 0.2 & 71.0 $\pm$ 0.1 & 67.7 $\pm$ 0.8 \\
GPT-OSS DiZiNER & 69.4 $\pm$ 0.4 & 69.4 $\pm$ 0.8 & \cellcolor{gray!15}70.1 $\pm$ 0.5 & 67.3 $\pm$ 0.8 & 66.2 $\pm$ 0.5 & 69.6 $\pm$ 0.2 & 67.7 $\pm$ 0.7 \\
Qwen DiZiNER & 69.4 $\pm$ 0.3 & 68.5 $\pm$ 0.3 & \cellcolor{gray!15}71.0 $\pm$ 0.1 & 66.1 $\pm$ 0.2 & 66.0 $\pm$ 0.3 & 69.2 $\pm$ 0.1 & 69.8 $\pm$ 0.7 \\
Average & 70.0 $\pm$ 0.9 & 68.3 $\pm$ 1.0 & \cellcolor{gray!15}71.4 $\pm$ 1.2 & 67.1 $\pm$ 0.7 & 66.3 $\pm$ 0.4 & 69.9 $\pm$ 0.8 & 68.4 $\pm$ 1.0 \\
\midrule
\multicolumn{8}{l}{\textit{NoiseBench}} \\
GPT-OSS Schema & 84.4 $\pm$ 0.4 & 84.2 $\pm$ 1.2 & 83.1 $\pm$ 0.5 & \cellcolor{gray!15}86.6 $\pm$ 0.4 & 84.2 $\pm$ 0.4 & 85.9 $\pm$ 0.5 & 83.7 $\pm$ 0.5 \\
GPT-OSS EvoPrompt & 75.4 $\pm$ 0.1 & 74.6 $\pm$ 1.4 & 75.6 $\pm$ 0.1 & 74.5 $\pm$ 0.9 & \cellcolor{gray!15}75.8 $\pm$ 0.7 & 75.3 $\pm$ 1.4 & 75.2 $\pm$ 0.6 \\
GPT-OSS Basic & 71.6 $\pm$ 0.2 & 73.1 $\pm$ 0.8 & 70.4 $\pm$ 0.8 & 73.9 $\pm$ 0.5 & \cellcolor{gray!15}74.9 $\pm$ 0.2 & 72.2 $\pm$ 0.7 & 71.1 $\pm$ 0.5 \\
Average & 77.1 $\pm$ 5.4 & 77.3 $\pm$ 4.9 & 76.4 $\pm$ 5.2 & \cellcolor{gray!15}78.3 $\pm$ 5.8 & 78.3 $\pm$ 4.2 & 77.8 $\pm$ 5.9 & 76.6 $\pm$ 5.2 \\
\midrule
\multicolumn{8}{l}{\textit{OntoNotes}} \\
GPT-OSS EvoPrompt & 65.5 $\pm$ 0.2 & 56.8 $\pm$ 5.5 & 65.6 $\pm$ 0.1 & 64.0 $\pm$ 0.6 & 64.3 $\pm$ 0.2 & 65.2 $\pm$ 0.3 & \cellcolor{gray!15}65.7 $\pm$ 0.2 \\
GPT-OSS Basic & 60.8 $\pm$ 0.2 & 49.0 $\pm$ 3.3 & \cellcolor{gray!15}61.8 $\pm$ 0.4 & 59.6 $\pm$ 0.5 & 61.0 $\pm$ 0.2 & 60.4 $\pm$ 0.6 & 61.7 $\pm$ 0.1 \\
Qwen DiZiNER & 61.4 $\pm$ 0.2 & 51.6 $\pm$ 1.2 & \cellcolor{gray!15}63.9 $\pm$ 0.1 & 60.1 $\pm$ 1.1 & 59.3 $\pm$ 0.2 & 61.3 $\pm$ 0.2 & 63.9 $\pm$ 0.2 \\
Average & 62.5 $\pm$ 2.1 & 52.5 $\pm$ 3.2 & \cellcolor{gray!15}63.8 $\pm$ 1.5 & 61.2 $\pm$ 2.0 & 61.6 $\pm$ 2.1 & 62.3 $\pm$ 2.1 & 63.7 $\pm$ 1.6 \\
\midrule
\multicolumn{8}{l}{\textit{Wikigold}} \\
GPT-OSS Schema & 70.3 $\pm$ 1.8 & 65.4 $\pm$ 9.6 & \cellcolor{gray!15}72.9 $\pm$ 1.6 & 61.1 $\pm$ 10.0 & 71.1 $\pm$ 2.4 & 67.8 $\pm$ 0.7 & 67.8 $\pm$ 4.2 \\
GPT-OSS EvoPrompt & 66.5 $\pm$ 0.2 & 67.5 $\pm$ 1.6 & \cellcolor{gray!15}67.9 $\pm$ 1.4 & 65.7 $\pm$ 0.9 & 64.7 $\pm$ 2.5 & 65.6 $\pm$ 1.1 & 66.6 $\pm$ 1.7 \\
Qwen DiZiNER & 65.0 $\pm$ 0.4 & 57.6 $\pm$ 4.7 & \cellcolor{gray!15}65.8 $\pm$ 0.7 & 59.7 $\pm$ 5.8 & 62.5 $\pm$ 1.8 & 63.2 $\pm$ 1.9 & 62.9 $\pm$ 1.7 \\
Average & 67.3 $\pm$ 2.2 & 63.5 $\pm$ 4.3 & \cellcolor{gray!15}68.9 $\pm$ 3.0 & 62.2 $\pm$ 2.6 & 66.1 $\pm$ 3.7 & 65.5 $\pm$ 1.9 & 65.8 $\pm$ 2.1 \\
\bottomrule
\end{tabular}
\caption{Test F1s of error-type-aware losses on LLM label variants, compared with cross-entropy (CE) and global masking ($L_{\text{all}}$), using \texttt{XLM-RoBERTa (large)}.}
\label{tab:xlm-roberta-large-test-f1-wide}
\end{table*}

%% file: tables/exp2_comparison_to_other_losses_xlm-roberta.tex
\begin{table*}[t]
\centering
\small
\setlength{\tabcolsep}{3pt}
\begin{tabular}{lcccc}
\toprule
Loss & CDR & NoiseBench & Ontonotes & Wikigold \\
\midrule
\multicolumn{3}{l}{\textit{Baseline losses}} \\
CE & 70.0 $\pm$ 0.9 & 77.1 $\pm$ 5.4 & 62.5 $\pm$ 2.1 & 67.3 $\pm$ 2.2 \\
BMM bootstrap & 69.9 $\pm$ 0.8 & 77.1 $\pm$ 5.5 & 62.6 $\pm$ 1.9 & 66.9 $\pm$ 2.5 \\
Corrected NLL & 69.1 $\pm$ 1.3 & 76.9 $\pm$ 5.4 & 62.6 $\pm$ 2.1 & 66.7 $\pm$ 3.7 \\
GCE & 69.5 $\pm$ 0.6 & 76.6 $\pm$ 5.0 & 62.3 $\pm$ 1.9 & 65.5 $\pm$ 2.7 \\
Focal & 70.1 $\pm$ 0.8 & 77.1 $\pm$ 5.4 & 62.6 $\pm$ 1.8 & 66.0 $\pm$ 2.6 \\
$L_{\text{all}}$ & 68.3 $\pm$ 1.0 & 77.3 $\pm$ 4.9 & 52.5 $\pm$ 3.2 & 63.5 $\pm$ 4.3 \\
\midrule
\multicolumn{3}{l}{\textit{Error-type-aware losses}} \\
$L_{\text{missing}}$ & \cellcolor{gray!15}71.4 $\pm$ 1.2 & 76.4 $\pm$ 5.2 & \cellcolor{gray!15}63.8 $\pm$ 1.5 & \cellcolor{gray!15}68.9 $\pm$ 3.0 \\
$L_{\text{entity}}$ & 67.1 $\pm$ 0.7 & \cellcolor{gray!15}78.3 $\pm$ 5.8 & 61.2 $\pm$ 2.0 & 62.2 $\pm$ 2.6 \\
$L_{\text{FP}}$ & 66.3 $\pm$ 0.4 & 78.3 $\pm$ 4.2 & 61.6 $\pm$ 2.1 & 66.1 $\pm$ 3.7 \\
$L_{\text{type}}$ & 69.9 $\pm$ 0.8 & 77.8 $\pm$ 5.9 & 62.3 $\pm$ 2.1 & 65.5 $\pm$ 1.9 \\
$L_{\text{missing, entity}}$ & 68.4 $\pm$ 1.0 & 76.6 $\pm$ 5.2 & 63.7 $\pm$ 1.6 & 65.8 $\pm$ 2.1 \\
\bottomrule
\end{tabular}
\caption{Comparison of error-aware losses with other related noise-robust losses w.r.t. test F1, using \texttt{XLM-RoBERTa (large)}.}
\label{tab:xlm-roberta-large-test-f1-dataset-averages-wide}
\end{table*}

%% file: tables/exp2_table_other_noises_distilbert.tex
\begin{table*}[t]
\centering
\small
\setlength{\tabcolsep}{3pt}
\begin{tabular}{lccccccc}
\toprule
Label variant & CE & $L_{\text{all}}$ & $L_{\text{missing}}$ & $L_{\text{entity}}$ & $L_{\text{FP}}$ & $L_{\text{type}}$ & $L_{\text{missing,entity}}$ \\
\midrule
\multicolumn{8}{l}{\textit{BC5CDR}} \\
Clean & \cellcolor{gray!15}83.6 $\pm$ 0.1 & 74.2 $\pm$ 1.5 & 83.3 $\pm$ 0.0 & 82.3 $\pm$ 0.3 & 82.9 $\pm$ 0.2 & 83.6 $\pm$ 0.1 & 83.3 $\pm$ 0.0 \\
Distant & 71.2 $\pm$ 0.3 & 62.1 $\pm$ 4.7 & 72.2 $\pm$ 0.3 & 67.2 $\pm$ 0.6 & 67.0 $\pm$ 0.6 & 71.6 $\pm$ 0.1 & \cellcolor{gray!15}72.2 $\pm$ 0.3 \\
\midrule
\multicolumn{8}{l}{\textit{NoiseBench}} \\
Clean & \cellcolor{gray!15}89.9 $\pm$ 0.2 & 84.8 $\pm$ 0.2 & 89.3 $\pm$ 0.3 & 87.9 $\pm$ 0.5 & 89.3 $\pm$ 0.2 & 88.6 $\pm$ 0.4 & 89.3 $\pm$ 0.3 \\
Expert & 85.9 $\pm$ 0.2 & 84.2 $\pm$ 2.1 & 86.4 $\pm$ 0.1 & 85.0 $\pm$ 0.6 & 85.5 $\pm$ 0.2 & 85.5 $\pm$ 0.5 & \cellcolor{gray!15}86.4 $\pm$ 0.1 \\
Crowd++ & 81.8 $\pm$ 0.3 & 79.4 $\pm$ 2.2 & 82.5 $\pm$ 0.1 & 79.7 $\pm$ 0.2 & 80.5 $\pm$ 0.5 & 80.9 $\pm$ 0.4 & \cellcolor{gray!15}82.5 $\pm$ 0.1 \\
Crowd & 64.4 $\pm$ 0.7 & 61.0 $\pm$ 4.8 & 66.1 $\pm$ 0.7 & 63.7 $\pm$ 0.4 & 64.2 $\pm$ 1.0 & 64.3 $\pm$ 1.4 & \cellcolor{gray!15}66.1 $\pm$ 0.7 \\
Distant & 65.1 $\pm$ 1.1 & 64.5 $\pm$ 2.3 & 66.3 $\pm$ 0.9 & 63.3 $\pm$ 1.1 & 64.6 $\pm$ 0.1 & 64.4 $\pm$ 0.8 & \cellcolor{gray!15}66.3 $\pm$ 0.9 \\
Weak & 60.0 $\pm$ 0.1 & 61.8 $\pm$ 2.2 & 60.1 $\pm$ 0.2 & 63.8 $\pm$ 1.0 & \cellcolor{gray!15}63.8 $\pm$ 0.3 & 61.3 $\pm$ 0.4 & 60.1 $\pm$ 0.2 \\
GPT3.5 & 55.5 $\pm$ 0.2 & 56.8 $\pm$ 1.4 & 55.1 $\pm$ 0.4 & \cellcolor{gray!15}60.4 $\pm$ 0.9 & 59.5 $\pm$ 0.4 & 56.0 $\pm$ 0.3 & 55.1 $\pm$ 0.4 \\
\midrule
\multicolumn{8}{l}{\textit{OntoNotes}} \\
Clean & \cellcolor{gray!15}76.1 $\pm$ 0.2 & 70.5 $\pm$ 0.9 & 75.9 $\pm$ 0.5 & 74.0 $\pm$ 0.4 & 74.8 $\pm$ 0.4 & 75.6 $\pm$ 0.2 & 75.9 $\pm$ 0.5 \\
Distant & 64.4 $\pm$ 0.6 & 62.1 $\pm$ 2.7 & \cellcolor{gray!15}64.5 $\pm$ 0.3 & 62.0 $\pm$ 0.4 & 63.2 $\pm$ 0.7 & 63.8 $\pm$ 0.2 & 59.9 $\pm$ 0.7 \\
\midrule
\multicolumn{8}{l}{\textit{Wikigold}} \\
Clean & \cellcolor{gray!15}73.4 $\pm$ 0.4 & 69.2 $\pm$ 2.1 & 71.8 $\pm$ 1.3 & 70.9 $\pm$ 1.8 & 72.6 $\pm$ 0.5 & 73.0 $\pm$ 2.2 & 71.8 $\pm$ 1.3 \\
Distant & 56.4 $\pm$ 0.3 & \cellcolor{gray!15}63.7 $\pm$ 1.3 & 57.9 $\pm$ 1.1 & 61.1 $\pm$ 3.5 & 62.1 $\pm$ 1.2 & 56.7 $\pm$ 1.1 & 57.9 $\pm$ 1.1 \\
\bottomrule
\end{tabular}
\caption{Test F1s of error-type-aware losses on extended label variants, compared with cross-entropy (CE) and global masking ($L_{\text{all}}$), using \texttt{DistilBERT}.}
\label{tab:distilbert-other-noises}
\end{table*}

%% file: tables/exp2_table_other_noises_xlm-roberta.tex
\begin{table*}[t]
\centering
\small
\setlength{\tabcolsep}{3pt}
\begin{tabular}{lccccccc}
\toprule
Label variant & CE & $L_{\text{all}}$ & $L_{\text{missing}}$ & $L_{\text{entity}}$ & $L_{\text{FP}}$ & $L_{\text{type}}$ & $L_{\text{missing,entity}}$ \\
\midrule
\multicolumn{8}{l}{\textit{BC5CDR}} \\
Clean & \cellcolor{gray!15}85.4 $\pm$ 0.1 & 79.9 $\pm$ 0.9 & 84.6 $\pm$ 0.2 & 82.4 $\pm$ 0.6 & 82.2 $\pm$ 0.3 & 85.4 $\pm$ 0.1 & 84.6 $\pm$ 0.1 \\
Distant & 73.9 $\pm$ 0.6 & 70.5 $\pm$ 1.1 & 77.2 $\pm$ 0.2 & 64.2 $\pm$ 3.7 & 67.2 $\pm$ 0.3 & 74.0 $\pm$ 0.1 & \cellcolor{gray!15}77.8 $\pm$ 0.4 \\
\midrule
\multicolumn{8}{l}{\textit{NoiseBench}} \\
Clean & \cellcolor{gray!15}94.8 $\pm$ 0.2 & 90.4 $\pm$ 1.8 & 94.8 $\pm$ 0.4 & 91.1 $\pm$ 0.4 & 93.8 $\pm$ 1.1 & 62.9 $\pm$ 44.5 & 94.6 $\pm$ 0.1 \\
Expert & \cellcolor{gray!15}90.2 $\pm$ 0.2 & 87.0 $\pm$ 0.7 & 60.2 $\pm$ 42.5 & 87.7 $\pm$ 1.0 & 88.8 $\pm$ 0.6 & 89.8 $\pm$ 0.5 & 87.7 $\pm$ 1.3 \\
Crowd++ & 85.1 $\pm$ 0.4 & 81.8 $\pm$ 1.4 & \cellcolor{gray!15}86.8 $\pm$ 0.2 & 80.0 $\pm$ 3.1 & 83.6 $\pm$ 1.6 & 84.9 $\pm$ 1.1 & 82.5 $\pm$ 1.0 \\
Crowd & 67.3 $\pm$ 0.6 & 67.3 $\pm$ 1.5 & \cellcolor{gray!15}70.4 $\pm$ 1.1 & 67.4 $\pm$ 1.5 & 56.1 $\pm$ 2.3 & 66.7 $\pm$ 1.1 & 68.4 $\pm$ 6.1 \\
Distant & 67.7 $\pm$ 0.8 & \cellcolor{gray!15}69.8 $\pm$ 1.5 & 69.7 $\pm$ 0.4 & 58.7 $\pm$ 7.9 & 64.8 $\pm$ 0.3 & 68.1 $\pm$ 0.7 & 47.7 $\pm$ 31.5 \\
Weak & 61.4 $\pm$ 0.9 & 44.2 $\pm$ 31.3 & 61.2 $\pm$ 0.4 & \cellcolor{gray!15}66.2 $\pm$ 0.8 & 64.4 $\pm$ 0.4 & 63.2 $\pm$ 0.9 & 42.9 $\pm$ 30.4 \\
GPT3.5 & 57.6 $\pm$ 0.3 & 59.1 $\pm$ 4.1 & 57.3 $\pm$ 0.9 & 60.5 $\pm$ 0.4 & \cellcolor{gray!15}62.0 $\pm$ 1.0 & 58.7 $\pm$ 0.4 & 39.6 $\pm$ 28.0 \\
\midrule
\multicolumn{8}{l}{\textit{OntoNotes}} \\
Clean & \cellcolor{gray!15}82.0 $\pm$ 0.0 & 56.4 $\pm$ 1.0 & 81.9 $\pm$ 0.2 & 80.2 $\pm$ 0.3 & 80.3 $\pm$ 0.4 & 81.7 $\pm$ 0.2 & 81.9 $\pm$ 0.2 \\
Distant & 69.6 $\pm$ 0.1 & 41.6 $\pm$ 6.1 & 69.9 $\pm$ 0.5 & 68.5 $\pm$ 0.4 & 68.5 $\pm$ 0.4 & 69.9 $\pm$ 0.1 & \cellcolor{gray!15}69.9 $\pm$ 0.5 \\
\midrule
\multicolumn{8}{l}{\textit{Wikigold}} \\
Clean & 79.7 $\pm$ 2.3 & 78.0 $\pm$ 3.5 & 78.0 $\pm$ 1.3 & 80.0 $\pm$ 1.4 & 79.5 $\pm$ 1.5 & \cellcolor{gray!15}80.2 $\pm$ 2.4 & 78.0 $\pm$ 1.3 \\
Distant & 64.7 $\pm$ 1.1 & \cellcolor{gray!15}69.7 $\pm$ 3.2 & 67.8 $\pm$ 0.5 & 60.9 $\pm$ 4.8 & 68.0 $\pm$ 1.2 & 61.6 $\pm$ 2.5 & 66.6 $\pm$ 1.7 \\
\bottomrule
\end{tabular}
\caption{Test F1s of error-type-aware losses on extended label variants, compared with cross-entropy (CE) and global masking ($L_{\text{all}}$), using \texttt{XLM-RoBERTa (large)}. }
\label{tab:xlm-roberta-other-noises}
\end{table*}